\documentclass[11pt]{article}
\PassOptionsToPackage{table}{xcolor}

\usepackage[final]{acl}

\usepackage{times}
\usepackage{latexsym}

\usepackage[T1]{fontenc}
\usepackage[utf8]{inputenc}

\usepackage{microtype}

\usepackage{inconsolata}

\usepackage{graphicx}

\usepackage{amssymb}
\usepackage{algorithm}
\usepackage{algpseudocode}
\makeatletter
\renewcommand{\fnum@algorithm}{\textbf{Algorithm~\thealgorithm:}}
\makeatother

\usepackage{makecell}
\usepackage{amsmath}
\floatname{algorithm}{Algorithm}

\usepackage{booktabs} % For professional looking tables
\usepackage{multirow}
\usepackage{enumitem}

\title{Alloc-MoE: Budget-Aware Expert Activation Allocation for Efficient Mixture-of-Experts Inference}

\author{
Baihui Liu, Kaiyuan Tian, Wei Wang, Zhaoning Zhang\thanks{Corresponding Author.}, Linbo Qiao, Dongsheng Li \\
National Key Laboratory of Parallel and Distributed Computing,\\ 
College of Computer Science and Technology,\\ 
National University of Defense Technology\\
\texttt{\{lbh,kyt,wwking,zhangzhaoning,qiao.linbo,dsli\}@nudt.edu.cn}}

\begin{document}
\maketitle

\begin{abstract}
Mixture-of-Experts (MoE) has become a dominant architecture for scaling large language models due to their sparse activation mechanism. 
However, the substantial number of expert activations creates a critical latency bottleneck during inference, especially in resource-constrained deployment scenarios. Existing approaches that reduce expert activations potentially lead to severe model performance degradation.
In this work, we introduce the concept of \emph{activation budget} as a constraint on the number of expert activations and propose Alloc-MoE, a unified framework that optimizes budget allocation coordinately at both the layer and token levels to minimize performance degradation. At the layer level, we introduce Alloc-L, which leverages sensitivity profiling and dynamic programming to determine the optimal allocation of expert activations across layers. At the token level, we propose Alloc-T, which dynamically redistributes activations based on routing scores, optimizing budget allocation without increasing latency. 
Extensive experiments across multiple MoE models demonstrate that Alloc-MoE maintains model performance under a constrained activation budget. Especially, Alloc-MoE achieves $1.15\times$ prefill and $1.34\times$ decode speedups on DeepSeek-V2-Lite at half of the original budget.
\end{abstract}

\begin{figure}[t]
\centering
\begin{minipage}{0.45\linewidth}
  \centering
  \includegraphics[width=\linewidth]{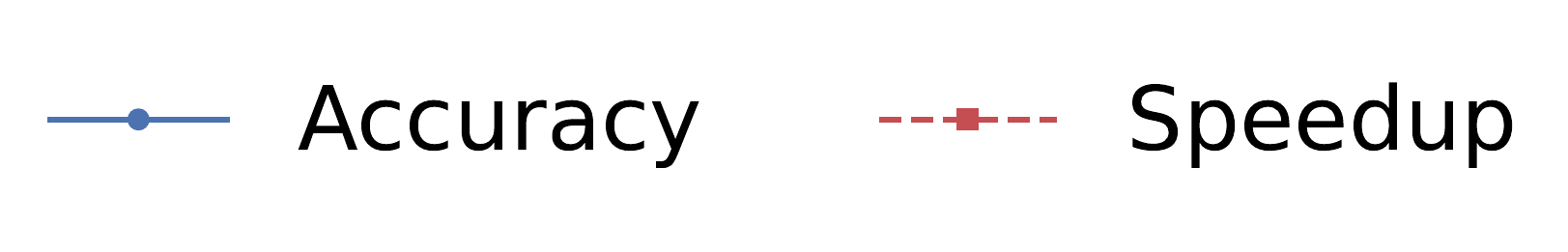}
  % \vspace{-4mm}
\end{minipage}\hfill
\begin{minipage}{0.53\linewidth}
  \centering
  \includegraphics[width=\linewidth]{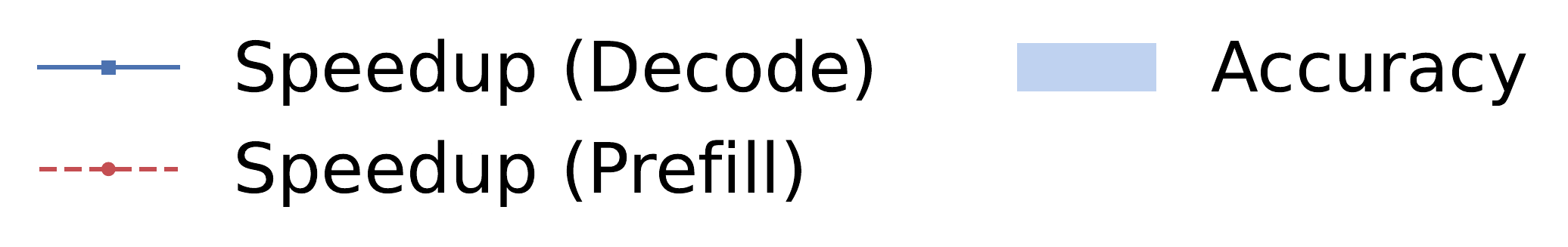}
  % \vspace{-1mm}
\end{minipage}\hfill

% \vspace{-0mm}

% ---------- (a) ----------
\begin{minipage}{0.5\linewidth}
  \centering
  \includegraphics[width=\linewidth]{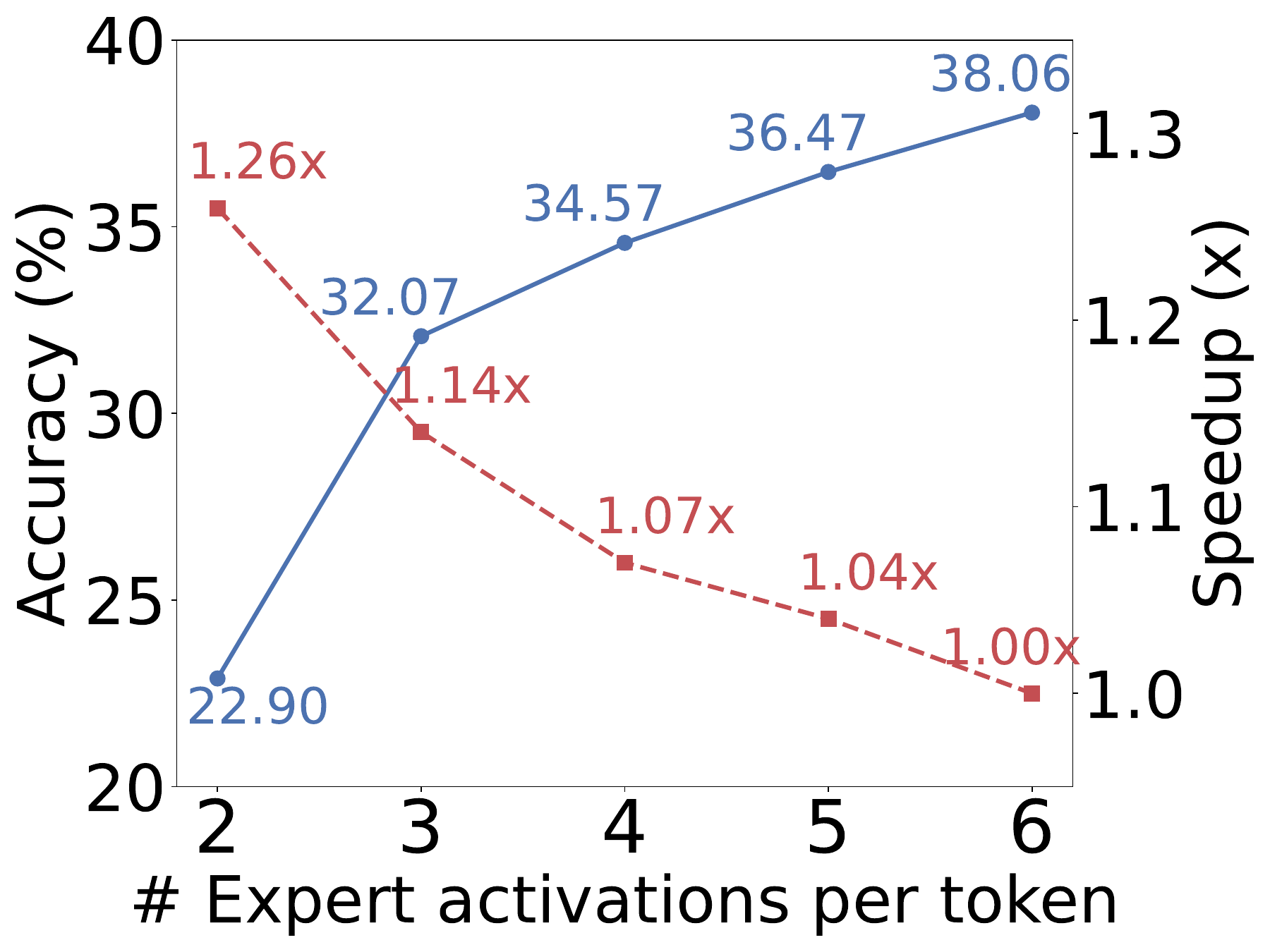}
  \vspace{-2mm}
  \small {\hspace{0em}(a)}
\end{minipage}\hfill
% ---------- (b) ----------
\begin{minipage}{0.5\linewidth}
  \centering
  \includegraphics[width=\linewidth]{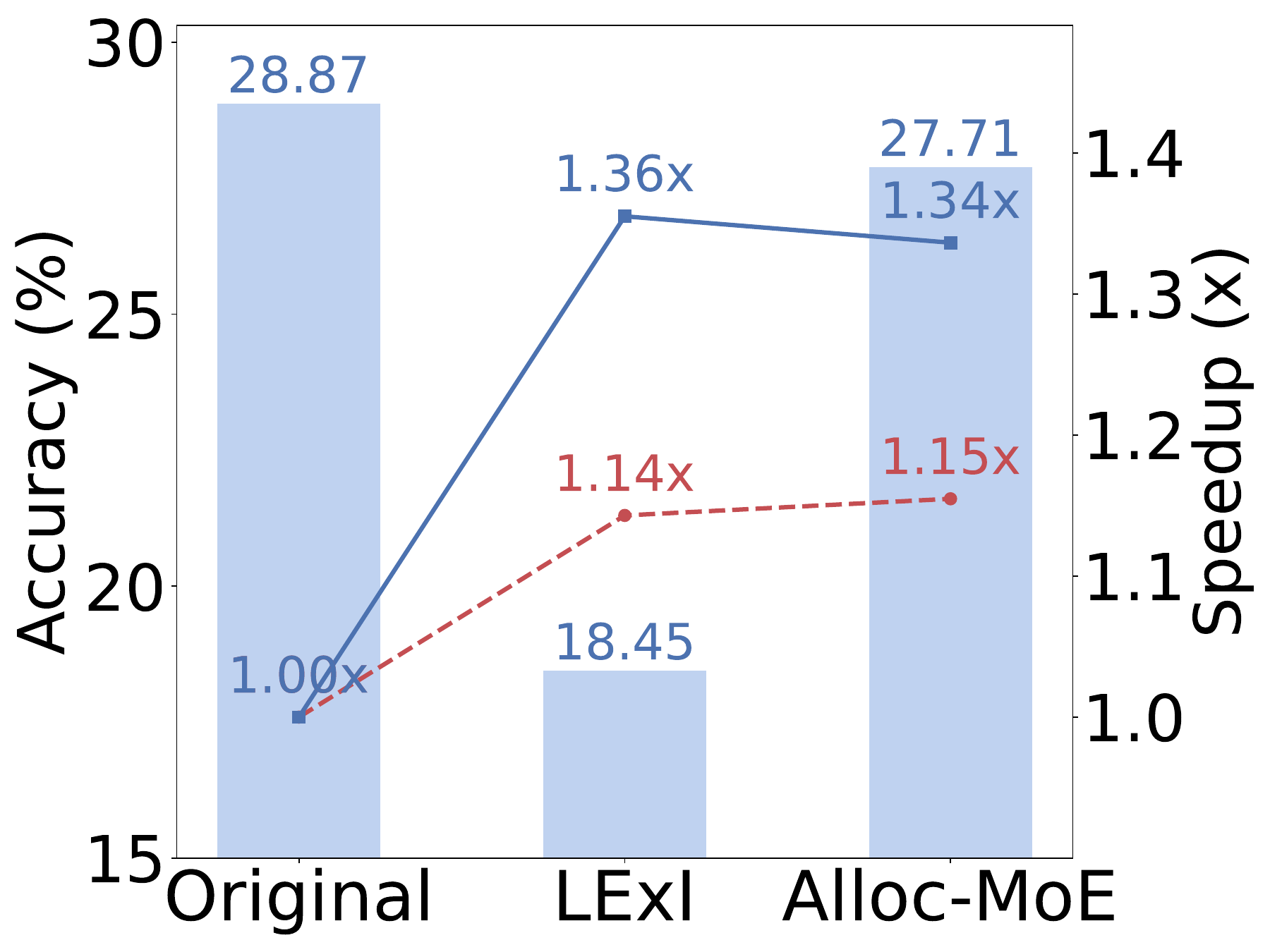}
  \vspace{-2mm}
  \small {\hspace{0em}(b)}
\end{minipage}\hfill

\caption{
(a) Reducing the number of expert activations per token increases speedup but decreases accuracy. 
(b) Alloc-MoE achieves inference latency comparable to the mainstream expert activation reduction method while retaining performance close to that of the original full-activation model when halving expert activation per token.
All results are measured on DeepSeek-V2-Lite.
Speedup is computed as the inference latency of the original model (with 6 expert activations per token) divided by the latency of the evaluated method.
}
\label{fig:insight and my trade-off}
\end{figure}

\section{Introduction}

Mixture-of-Experts (MoE) has emerged as an important approach for sparsifying mainstream Transformer-based models~(\citealp{SMoE, Gshard, SwitchTransformers}). It replaces the feed-forward network layers with sparse MoE layers, which consist of a gating network and a set of small networks named \emph{experts}. The gating network computes routing scores over experts per token and activates only the Top-K experts for each token. This sparse expert activation mechanism facilitates widespread deployment of MoE models in real-world systems~(\citealp{deepseekai2024deepseekv3technicalreport,qwen3techreport,kimiteam2025kimik2openagentic}).

However, as the number of input tokens increases, the large amount of expert activations becomes a critical bottleneck for efficient MoE inference, which is further severe in resource-constrained deployment scenarios. Existing works have explored reducing expert activations either from token-level~\cite{XMoE, DynamicMoE, DynMoE, DAMoE, NAEE, SeerMoE, AdapMoE, MCMoE} or layer-level~\cite{yang2025fastermoellminference, LExI} perspective to decrease inference latency. However, these approaches neglect their impact on model performance, potentially leading to significant degradation.
Figure~\ref{fig:insight and my trade-off}(a) shows that on DeepSeek-V2-Lite, reducing the expert activations per token from 6 to 3 leads to a 17\% performance degradation, which exacerbates to nearly 40\% when further reduced to two activated experts.

To address this problem, we formalize the number of expert activations as an \emph{activation budget}, which is closely correlated with inference latency, and propose \emph{Alloc-MoE}, a unified framework that optimizes budget allocation to minimize performance degradation under a fixed expert activation budget. 
Figure~\ref{fig:insight and my trade-off}(b) demonstrates that Alloc-MoE achieves a comparable speedup to the mainstream method and maintains performance close to the original model.
Alloc-MoE coordinately allocates the budget at the layer and token levels utilizing \emph{Alloc-L} and \emph{Alloc-T}, respectively.
Alloc-L adopts an end-to-end performance metric to profile layer sensitivity and formulates layer-level expert activation allocation as a sensitivity-guided optimization problem, which is solved exactly and efficiently via dynamic programming, yielding the optimal allocation of expert activations across layers.
Building upon this, Alloc-T dynamically redistributes expert activations across tokens within each layer according to token-level routing scores. By prioritizing tokens with less concentrated routing distributions, Alloc-T better allocates limited expert activations without introducing extra inference latency.

Our contributions are summarized as follows:
\begin{itemize}
    \item We introduce the expert activation budget and propose Alloc-MoE, a unified framework that optimizes the allocation of budgets coordinately at the layer and token levels.
    \item We present Alloc-L, a layer-level expert activation allocation method that optimizes layer-level expert activation allocation under a fixed budget by leveraging global sensitivity profiling and exact dynamic programming.
    \item We introduce Alloc-T, a token-level redistribution strategy that reallocates expert activations according to routing scores, improving model performance under a fixed activation budget without additional inference latency.
    \item Extensive experiments demonstrate that Alloc-MoE sustains performance under a restricted activation budget across multiple MoE models. Notably, on DeepSeek-V2-Lite, it attains $1.15\times$ speedup in prefill and $1.34\times$ in decode when using only half of the original budget.
\end{itemize}

\section{Related works}

\begin{figure*}[t]
  \includegraphics[width=\linewidth]{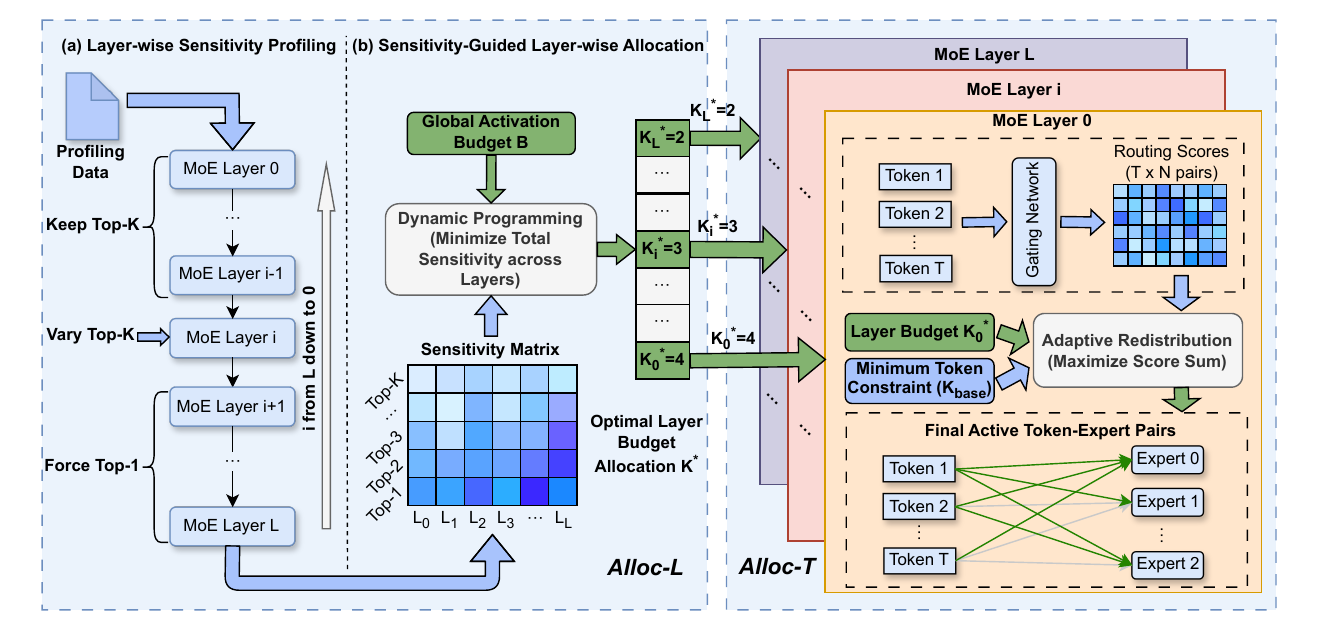}
  \caption{
    Overview of the Alloc-MoE framework, consisting of two components:
    \textbf{Alloc-L} (left) profiles layer-wise sensitivity to construct a sensitivity matrix and applies dynamic programming to determine the optimal layer-level activation budget $\mathbf{K}^*$ under a global expert activation budget $B$;
    \textbf{Alloc-T} (right) adaptively redistributes expert activations across tokens based on routing scores, maximizing the total routing weight while respecting the layer-wise budgets.
    }
  \label{fig:overview}
\end{figure*}

\paragraph{Token-level Expert Activation Reduction.}
These methods aim to adaptively reduce the number of activated experts per token based on token-level routing scores.
XMoE~\cite{XMoE} introduces Top-$P$ routing, where each token activates a variable number of experts whose cumulative routing scores exceed a predefined threshold $P$. However, it requires training-time calibration and can lead to over-activation.
Dynamic-MoE~\cite{DynamicMoE} further regularizes the entropy of routing score distributions during training to mitigate the over-activation behavior of Top-$P$ routing.
NAEE~\cite{NAEE} conditionally skips the secondary expert in Top-2 routing for Mixtral~\cite{jiang2024mixtralexperts} when its gating weight falls below a relative threshold, but this approach is limited to models with Top-2 routing and does not generalize to larger expert sets.
AdapMoE~\cite{AdapMoE} employs a sensitivity-aware gating mechanism with an offline-calibrated threshold to adaptively reduce expert activations while preserving model performance, yet it is similarly constrained to specific pretrained models and requires offline calibration.

\paragraph{Layer-level Expert Activation Allocation.}
These approaches typically reduce the number of per-layer expert activations and demonstrate the efficiency improvement at the cost of performance degradation.
\citet{yang2025fastermoellminference} investigate several heuristic expert reduction strategies, showing significant throughput improvements across both low- and high-concurrency settings, and revealing that their efficiency--accuracy trade-offs differ markedly across MoE models.
LExI~\cite{LExI} proposes a data-free, post-training layer-adaptive reduction method and demonstrates improved inference efficiency compared to pruning-based approaches.

\section{Methodology}
In this section, we first review the Top-K routing mechanism and formalize the notion of expert activation budget. Then we present the two key components of Alloc-MoE: Alloc-L for layer-level activation allocation and Alloc-T for token-level activation redistribution.

\subsection{Preliminary}
\paragraph{Top-K Routing of Mixture-of-Experts.}
A standard MoE layer consists of $N$ expert networks and a gating network that enables conditional computation by selectively activating a subset of experts for each token.
Given an input hidden representation $X \in \mathbb{R}^{T \times H}$, the gating network independently routes each token $x_t$ by projecting it with $W_{\text{gate}} \in \mathbb{R}^{H \times N}$ and computing routing scores over experts:
\begin{equation}
\label{eq:Top-K Softmax}
w_t = \text{Top-K}(\text{Softmax}(x_t \cdot W_{\text{gate}})),
\end{equation}
where only the $K$ largest scores are retained and the rest are masked to zero.
The MoE output is computed as a weighted sum over the activated experts,
\begin{equation}
\label{eq:weighted sum}
y_t = \sum_{i=1}^{N} w_{t,i} \cdot E_i(x_t),
\end{equation}
with $w_{t,i}=0$ for non-activated experts.

\paragraph{Expert Activation Budget.}
In an MoE layer, each activated expert corresponds to one feed-forward execution, typically implemented as a GEMM operation. As a result, the inference latency scales approximately linearly with the total number of expert activations.
We formalize expert activations using two closely related budget notions: a \emph{global activation budget} and a \emph{layer-level activation budget}.
Specifically, we define the \emph{global activation budget} as the total number of expert activations incurred by a single token across all MoE layers during inference. For a model with $L$ MoE layers, if a token activates up to Top-K experts in each layer, its global activation budget is bounded by $L \times K$.
We define the \emph{layer activation budget} as the average number of expert activations allocated per token within a given MoE layer. For a layer processing $T$ tokens with a Top-K routing, the total activation budget is at most $T \times K$, which can be flexibly redistributed across tokens as long as the average per-token budget is preserved.
The \emph{global activation budget} equals the sum of the \emph{layer activation budgets} across all layers.
This formalization enables principled allocation at both the layer and token levels, directly motivating the design of Alloc-L and Alloc-T.

\subsection{Alloc-L}
Alloc-L optimizes the allocation of expert activations across layers under a fixed global activation budget. 
It profiles layer-wise sensitivity using an end-to-end performance metric and leverages this information to perform a sensitivity-aware allocation, which can be solved efficiently via dynamic programming.

\paragraph{Layer-wise Sensitivity Profiling.}
The sensitivity of expert allocations in shallow layers may be masked by compensatory allocations in deeper layers, obscuring the attribution of observed performance changes to a specific layer. To better characterize the layer-wise sensitivity, we adopt an allocation-isolating profiling strategy that mitigates interference from subsequent layers.

Specifically, we utilize the perplexity metric. For a model with $L$ MoE layers indexed $\{0, 1,\dots, L-1\}$, when profiling target layer $i$, we gradually reduce its allocated Top-K value from the original $K_{\text{orig}}$ down to 1, while temporarily constraining all deeper layers to the minimal activation setting ($\text{Top-}K=1$) and keeping all preceding layers at the original $K_{\text{orig}}$.
This preserves routing patterns in preceding layers and mitigates compensatory effects from deeper layers, enabling accurate attribution of observed performance changes to the target layer.
The perplexity measured under different Top-K settings is then used to characterize the relative sensitivity of target layer $i$ to changes in expert activation.

After applying the profiling process in Algorithm~\ref{alg:SensitivityProfiling}, we obtain a global sensitivity matrix $S \in \mathbb{R}^{L \times K_{\text{orig}}}$, where each normalized row $S[i,1..K_{\text{orig}}]$ captures the relative loss of layer $i$ under varying expert activations and provides a globally comparable layer-wise sensitivity measure across layers.

\paragraph{Sensitivity-Guided Layer-wise Expert Allocation.}

Consider a model with $L$ MoE layers, let $B$ denote the maximum global budget, $K_{\text{orig}}$ the original Top-K of the model, and define $\mathbf{K} = [K_0, K_1, \dots, K_{L-1}]$ as a layer-level activation budget allocation, where $1\le K_i\le K_{\text{orig}}$ represents the budget of expert activations per token allocated to layer $i$.
Our goal is to identify the optimal layer-level allocation $\mathbf{K}^\ast$ that minimizes the aggregated loss under the global activation budget:
\begin{align}
    \underset{{{\mathbf{K}=[K_{i}]_{i=0}^{L-1}}}}{\operatorname{argmin}}\quad & \sum_{i=0}^{L-1} \mathbf{S}[i,K_i] \label{eq:search object} \\
    \text{s.t.} \quad & \sum_{i=0}^{L-1} K_i \leq B, \label{eq:constraints 1} \\
    & 1 \leq K_i \leq K_{\text{orig}},\quad \forall i. \label{eq:constraints 2}
\end{align}

We solve this problem by casting it as a budget-constrained allocation task analogous to a grouped knapsack problem. Each MoE layer constitutes a group of allocation choices, where activating $k$ experts per token for layer $i$ consumes a budget of $k$ and incurs a sensitivity cost of $\mathbf{S}[i,k]$. The objective is to minimize the total sensitivity across all layers under the global budget constraints.
We define $\mathrm{DP}[i,b]$ as the minimum cumulative sensitivity achievable by allocating experts to layer $0,1,\cdots,i$ under a total budget $b$.
The recurrence relation is:
\begin{equation}
  \label{eq:DP}
    \mathrm{DP}[i,b] = \min_{k \leq b}\ ( \mathrm{DP}[i-1,b - k] + \mathbf{S}[i,k])
\end{equation}
with base conditions $\mathrm{DP}[-1,0]=0$ and $\mathrm{DP}[-1,b>0]=+\infty$.

After processing all $L$ layers, the optimal allocation is obtained as $\mathbf{K}^\ast = \underset{b \le B}{\operatorname{argmin}}\   \mathrm{DP}[L-1,b]$, with the corresponding layer-wise expert allocation recovered via backtracking. This dynamic programming formulation yields an exact solution with complexity $O(L \cdot B \cdot K_{\text{orig}})$, which is efficient in practice due to the limited allocation range per layer, as the total budget $B$ is upper-bounded by $L \times K_{\text{orig}}$.

\begin{algorithm}[t]
\caption{Layer-wise Allocation-Isolated Sensitivity Profiling}
\label{alg:SensitivityProfiling}
\begin{algorithmic}[1]
\Require Model $M$, number of layers $L$, calibration dataset $D_{\text{calib}}$, original Top-K $K_{\text{orig}}$
\Ensure Global sensitivity matrix $S \in \mathbb{R}^{L \times K_{\text{orig}}}$

\State Initialize $S \gets 0$, $C \gets [K_{\text{orig}}]_{i=0}^{L-1}$
\State ApplyConfiguration($M, C$)
\State $PPL \gets$ GetPerplexity($M, D_{\text{calib}}$)

\For{$i = L-1$ \textbf{down to} $0$}
    \State $S[i,K_{\text{orig}}] \gets PPL$
    \For{$k = K_{\text{orig}}-1$ \textbf{down to} $1$}
        \State $C[i] \gets k$
        \State ApplyConfiguration($M, C$)
        \State $PPL \gets$ GetPerplexity($M, D_{\text{calib}}$)
        \State $S[i,k] \gets PPL$
    \EndFor
\EndFor

\State \Return $S$
\end{algorithmic}
\end{algorithm}

\subsection{Alloc-T}
Alloc-T optimizes the allocation of expert activations across tokens within each layer under a fixed layer-level activation budget. It leverages token-level routing score distributions to adaptively redistribute activations, improving allocation without increasing the overall inference cost.

\paragraph{Token-level Adaptive Expert Activation Redistribution.}

Alloc-T treats expert activation allocation within a layer as a collective allocation problem across tokens under a fixed layer-level activation budget, enabling expert activation to be adaptively allocated to tokens according to their routing score distributions.

Specifically, let $K_l$ denote the average activation budget per token for layer $l$ as determined by Alloc-L. 
We define the candidate set of token–expert pairs as
\begin{equation}
    \mathcal{C}_l = \{(t,e)| e \in \text{Top-K}(w_t), t=1,\dots,T\},
\end{equation}
where $T$ is the number of tokens and $|\mathcal{C}_l| = T \times K_l$. 
Alloc-T then selects activations from $\mathcal{C}_l$ while respecting the average activation budget $K_l$.

We introduce binary variables $z_{t,e}\in\{0,1\}$ to indicate whether expert $e$ is activated by token $t$, and formulate token-level expert activation allocation as:
\begin{align}
  \max_{{z_{t,e}}} \quad & \sum_{(t,e)\in\mathcal{C}_l} z_{t,e}\cdot w_{t,e} \label{eq:maximize object} \\
  \text{s.t.} \quad
  & \sum_{(t,e)\in\mathcal{C}_l} z_{t,e} \le T\times K_l, \label{eq:maximize constraint 1} \\
  & \sum_{e} z_{t,e} \ge K_{\text{base}}, \quad \forall t. \label{eq:maximize constraint 2}
\end{align}
Constraint~\ref{eq:maximize constraint 1} enforces the layer-level activation budget at layer $l$, while Constraint~\ref{eq:maximize constraint 2} ensures a minimum base expert activation allocation $K_{\text{base}}$ per token to prevent token dropping.
In practice, this constrained optimization is efficiently solved by using simple masking and global top-selection operations on the routing scores, incurring negligible inference overhead even at large $T$.

Specifically, given the routing score $\mathit{scores}$ of shape $[T, K_{\text{orig}}]$, each row $\mathit{scores}[i, :]$ contains the routing scores for token $i$, sorted in descending order. 
We first preserve the Top-$K_{\mathrm{base}}$ experts for each token to ensure routing stability and maintain a minimum allocation. Then, instead of performing independent per-token selection, we collect the remaining $K_{\text{orig}} - K_{\mathrm{base}}$ candidate expert scores across all tokens and globally select the top $(K_l - K_{\mathrm{base}}) \cdot T$ entries under the overall activation budget constraint. This global selection enables computation to be dynamically shifted toward tokens with less concentrated routing distributions.

Notably, standard Top-K routing is a special case of this formulation when $K_{\text{base}} = K_l$ and no additional budget is available for redistribution. 
Alloc-T therefore generalizes conventional routing by relaxing fixed expert activation allocation per token and enabling adaptive token-level expert activation allocation under a fixed layer-level activation budget.

\section{Experiments}

\subsection{Setup}
\paragraph{Models and Budgets.}
We evaluate Alloc-MoE on three representative MoE models: DeepSeek-V2-Lite~\cite{deepseekv2}, Qwen1.5-MoE-A2.7B~\cite{qwen_moe}, and OLMoE-1B-7B-0924~\cite{olmoe}. These models differ in scale, number of MoE layers and Top-K configurations, providing a diverse evaluation benchmark for studying expert activation allocation strategies. For brevity, we refer to them as DeepSeek, Qwen, and OLMoE respectively in the following experiments. Table~\ref{tab:model_specs} summarizes their architectures. 

For each model, we impose a strict global activation budget $B$.
Specifically, we evaluate DeepSeek with $B \in \{130, 104, 78, 52\}$, corresponding to average per-layer Top-K allocations of $\{5, 4, 3, 2\}$.
For Qwen, we consider $B \in \{84, 72, 60, 48\}$, corresponding to $\{3.5, 3, 2.5, 2\}$.
For OLMoE, we adopt $B \in \{112, 96, 80, 64\}$, corresponding to $\{7, 6, 5, 4\}$.
These budgets span mild to aggressive sparsification regimes, enabling systematic evaluation of performance–efficiency trade-offs under constrained expert activation allocation.

\begin{table}[t]
\centering
\setlength{\tabcolsep}{3pt}
\resizebox{0.9\linewidth}{!}{
\begin{tabular}{lccc}
\toprule
\textbf{Model} &
\textbf{\makecell{\# MoE\\Layers}} &
\textbf{\makecell{\# Act. / Tot.\\Experts}} &
\textbf{\makecell{\# Act. / Tot.\\Params.}} \\
\midrule
DeepSeek & 26 & 6 / 64 & 2.4B / 15.7B \\
Qwen & 24 & 4 / 60 & 2.7B / 14.3B \\
OLMoE & 16 & 8 / 64 & 1.0B / 7.0B \\
\bottomrule
\end{tabular}
}
\caption{Architectural details of the evaluated MoE models.}
\label{tab:model_specs}
\end{table}

\begin{figure*}[t]
\centering

% ---------- legend ----------
\begin{minipage}{\linewidth}
  \centering
  \includegraphics[width=0.9\linewidth]{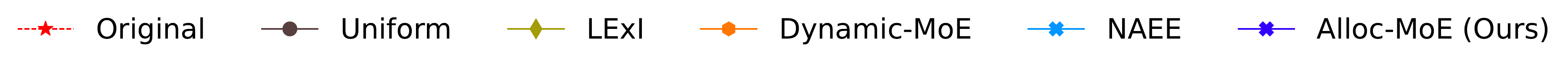}
\end{minipage}

\vspace{2mm}

% ---------- (a) ----------
\begin{minipage}{0.32\linewidth}
  \centering
  \includegraphics[width=\linewidth]{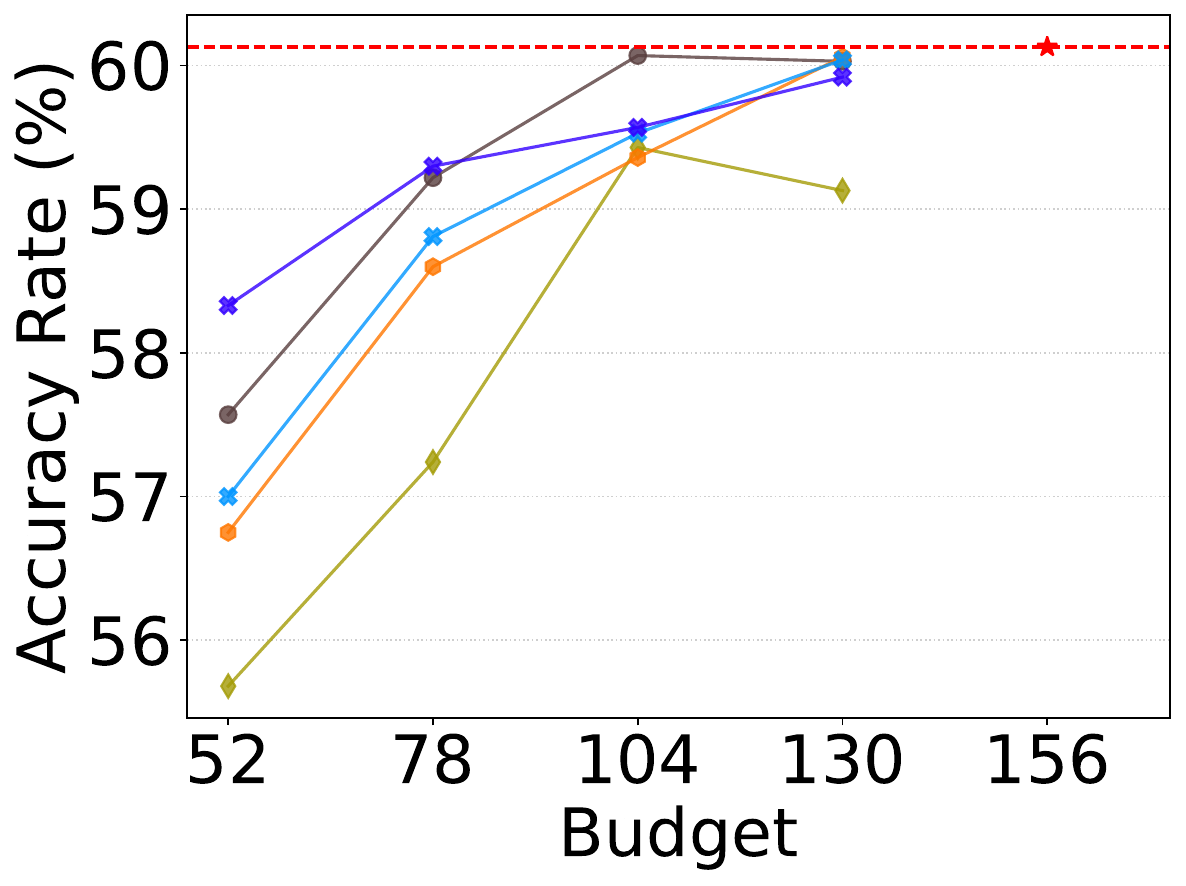}
  \vspace{-2mm}
  \small {\hspace{2em}(a)}
\end{minipage}\hfill
% ---------- (b) ----------
\begin{minipage}{0.32\linewidth}
  \centering
  \includegraphics[width=\linewidth]{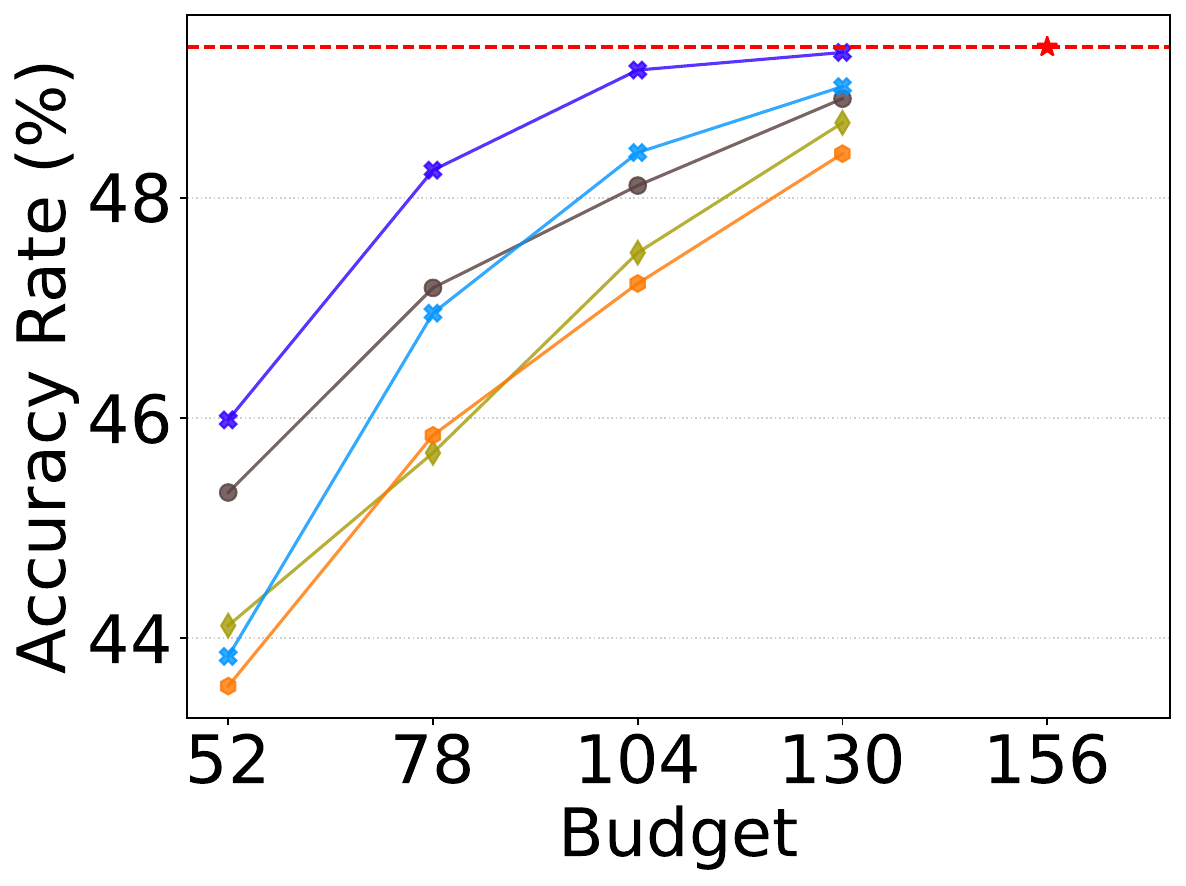}
  \vspace{-2mm}
  \small {\hspace{2em}(b)}
\end{minipage}\hfill
% ---------- (c) ----------
\begin{minipage}{0.32\linewidth}
  \centering
  \includegraphics[width=\linewidth]{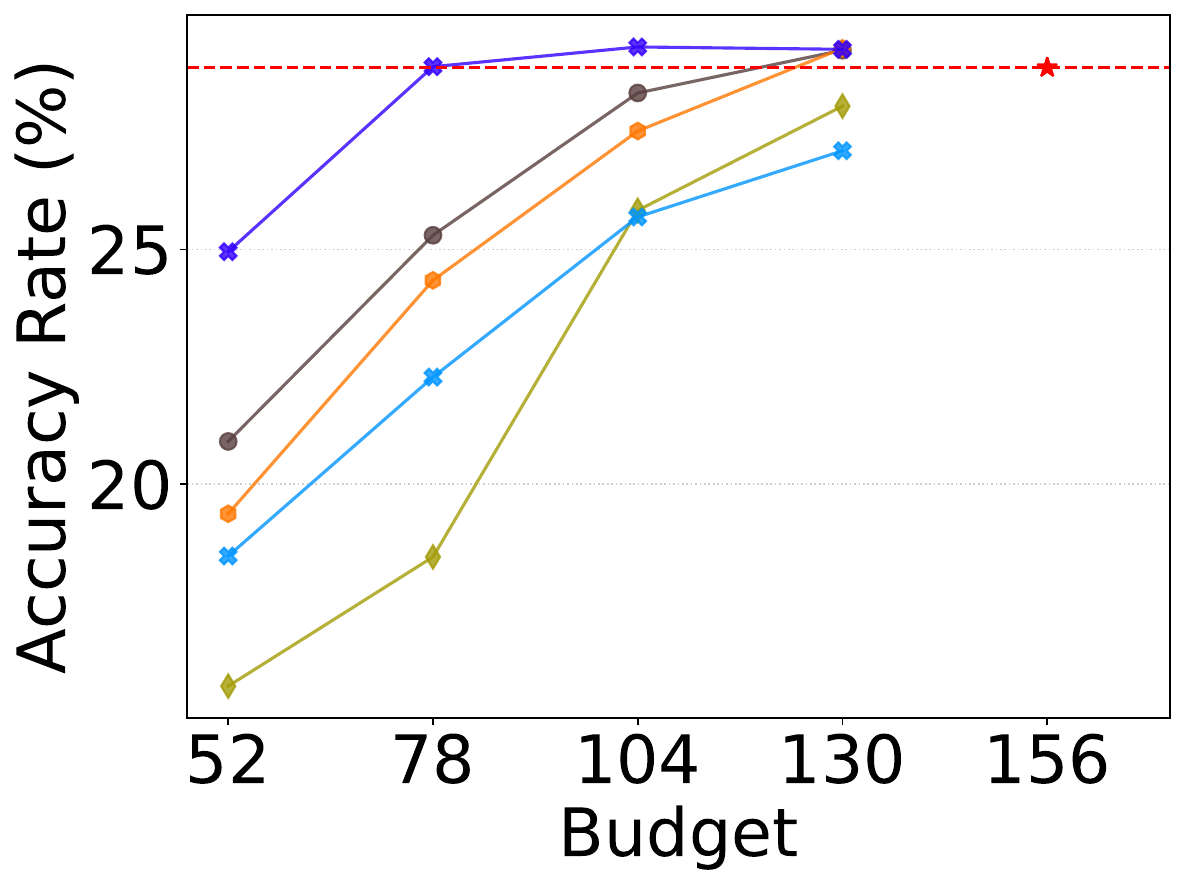}
  \vspace{-2mm}
  \small {\hspace{2em}(c)}
\end{minipage}
\caption{Alloc-MoE Results on DeepSeek-V2-Lite under varying global activation budgets. (a) NLU task, (b) Reasoning task and (c) Math task.}
\label{fig:DSV2 Alloc-MoE}
\end{figure*}

\paragraph{Baselines.}
We compare Alloc-MoE against representative MoE inference baselines that differ in how a fixed global activation budget is allocated across layers and tokens.
Specifically, we consider: 
\emph{Uniform}, which allocates identical expert activation to all MoE layers; 
\emph{LExI}, which redistributes the budget across layers based on intra-layer sensitivity profiling;
and two token-level baselines, \emph{Dynamic-MoE} and \emph{NAEE}, which adapt expert activation across tokens under the \emph{Uniform} layer-wise allocations.
% All methods are evaluated under identical expert activation budgets.
Implementation details of layer- and token-level baselines are provided in Appendix~\ref{subsec:heuristic_allocation}.

\paragraph{Datasets and Benchmarks.}
We use Wikitext2~\cite{wikitext} for calibration and conduct extensive evaluations on 20 datasets covering three task groups: Natural Language Understanding (NLU), Reasoning and Math.
The NLU benchmarks comprise BoolQ~\cite{clark-etal-2019-boolq}, LAMBADA~\cite{paperno-etal-2016-lambada}, RACE~\cite{lai-etal-2017-race}, SciQ~\cite{sciq}, MNLI, QNLI and RTE~\cite{wang-etal-2018-glue}.
The Reasoning tasks include ARC (ARC-E and ARC-C)~\cite{arc}, HellaSwag~\cite{zellers-etal-2019-hellaswag}, LogiQA~\cite{liu2021logiqa}, MMLU~\cite{hendryckstest2021mmlu}, PIQA~\cite{piqa}, TruthfulQA~\cite{lin-etal-2022-truthfulqa}, ACP~\cite{acp}, BBH~\cite{suzgun-etal-2023-challenging}, GroundedCocoa~\cite{kohli-etal-2025-groundcocoa}, and SWAG~\cite{zellers-etal-2018-swag}.
The Math benchmarks include GSM8K~\cite{cobbe2021gsm8k}, ASDiv~\cite{asdiv}, and MathQA~\cite{mathqa}.
For GSM8K, ACP, and BBH, we report exact match (EM) metric. Accuracy metric is used for all other benchmarks.
For clarity, we report the average performance within each task group.

\paragraph{Evaluation details.}
For performance evaluation, we use the \texttt{lm-eval}~\cite{eval-harness} framework with vllm~\cite{kwon2023efficient} backend. For inference efficiency evaluation, we report prefill and decode speedups relative to the original expert activation allocation strategy, using the DeepSeek model as a representative benchmark. \emph{LExI} allocation under the same global activation budgets is included as a reference. Implementation and measurement details are provided in Appendix~\ref{subsec:eval_details}. All experiments are conducted on a single NVIDIA H100 80GB GPU.

\begin{figure*}[t]
\centering

% ---------- legend ----------
\begin{minipage}{\linewidth}
  \centering
  \includegraphics[width=0.9\linewidth]{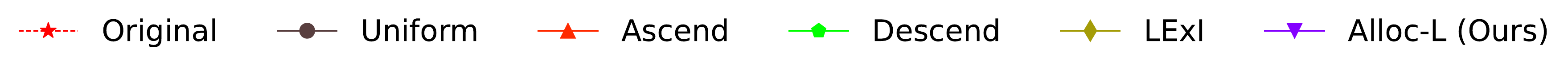}
\end{minipage}

\vspace{2mm}

% ---------- (a) ----------
\begin{minipage}{0.32\linewidth}
  \centering
  \includegraphics[width=\linewidth]{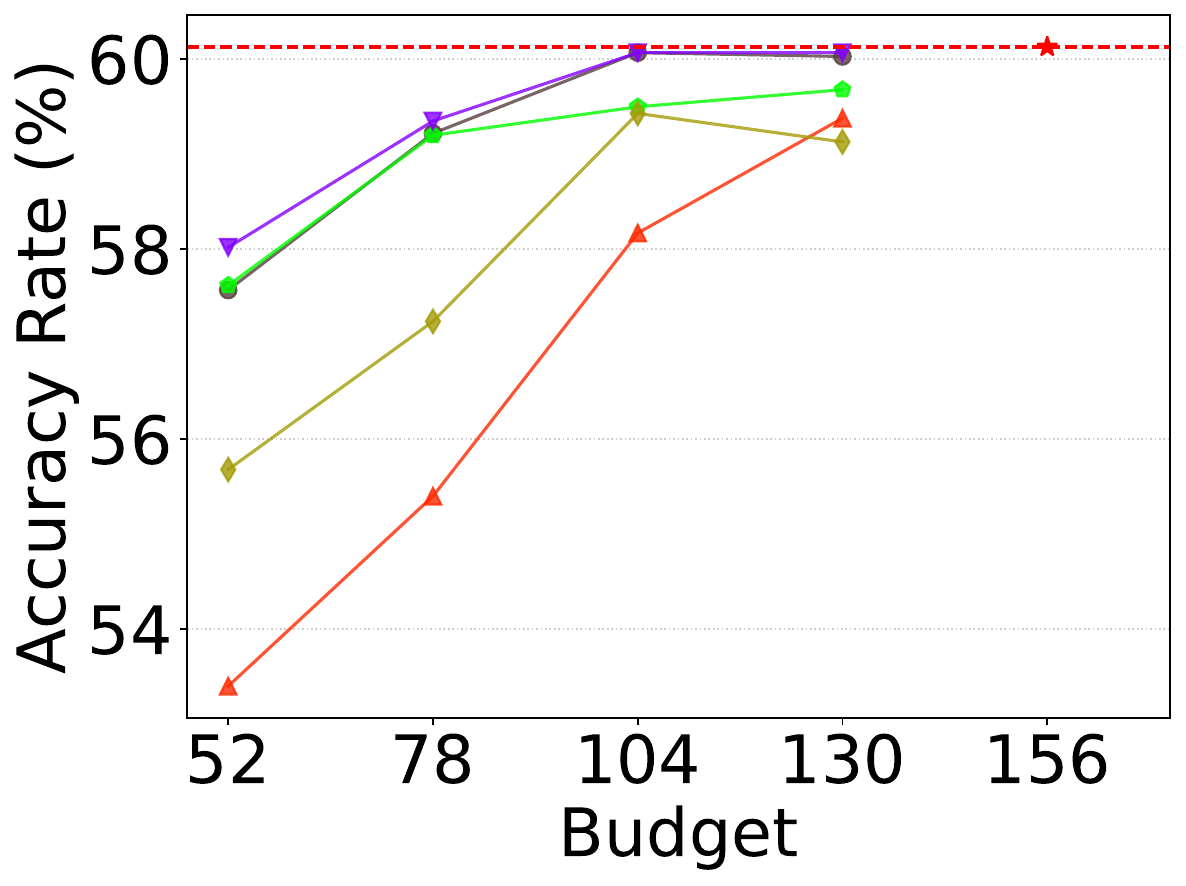}
  \vspace{-2mm}
  \small {\hspace{2em}(a)}
\end{minipage}\hfill
% ---------- (b) ----------
\begin{minipage}{0.32\linewidth}
  \centering
  \includegraphics[width=\linewidth]{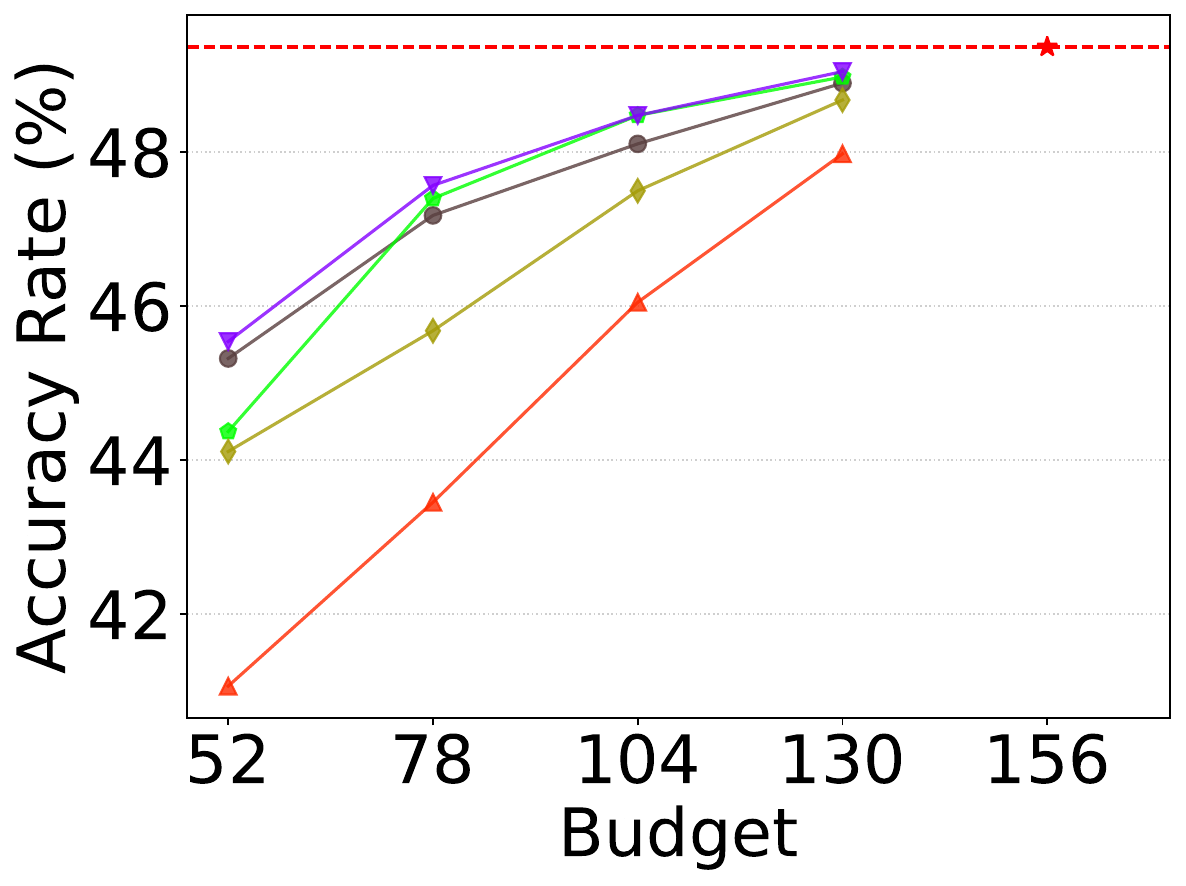}
  \vspace{-2mm}
  \small {\hspace{2em}(b)}
\end{minipage}\hfill
% ---------- (c) ----------
\begin{minipage}{0.32\linewidth}
  \centering
  \includegraphics[width=\linewidth]{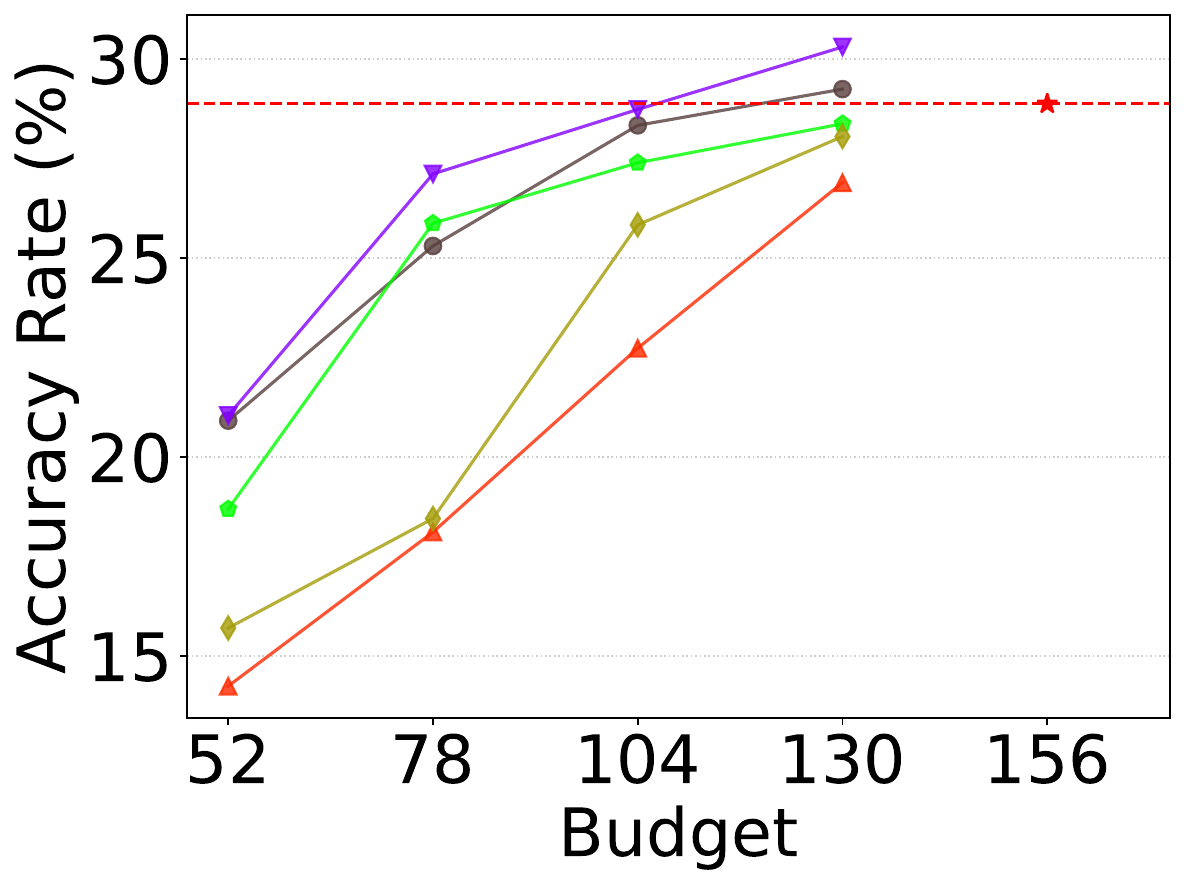}
  \vspace{-2mm}
  \small {\hspace{2em}(c)}
\end{minipage}

\caption{Ablation results of Alloc-L on DeepSeek-V2-Lite under varying global activation budgets. (a) shows NLU task, (b) Reasoning task and (c) Math task.}
\label{fig:DSV2 Alloc-L}
\end{figure*}

\subsection{Main Results}

\paragraph{Performance.}
Figure~\ref{fig:DSV2 Alloc-MoE} shows task-aggregated performance across varying budgets. 
From these results, several observations can be drawn:
(1) Alloc-MoE outperforms baselines in 10 of 12 evaluated settings, demonstrating robust generalization across tasks and budgets.
(2) As the budget becomes increasingly restrictive, Alloc-MoE exhibits minimal performance degradation relative to all baselines, underscoring its robustness to aggressive expert sparsification.
(3) The performance advantage of Alloc-MoE grows with task complexity, with average improvements of 0.05\% on NLU, 0.70\% on Reasoning, and 2.15\% on Math tasks. This indicates that Alloc-MoE is more beneficial for tasks with greater computational diversity, where adaptive activation allocation can better match the varying demands across tokens and layers, leading to more pronounced improvements compared to \emph{Uniform} allocation.
Similar trends are observed on Qwen and OLMoE in Appendix~\ref{subsec:additional_results} (Figure~\ref{fig:Qwen and OLMoE Alloc-MoE}), Alloc-MoE consistently maintains competitive performance under mild budget constraints and demonstrates increasingly clear advantages under aggressive sparsification, particularly on Reasoning and Math tasks.
Overall, these results validate Alloc-MoE as a robust and general expert activation allocation framework for MoE inference under constrained budgets.

\paragraph{Inference Efficiency.}

\begin{figure}[t]
\centering

% ---------- legend ----------
\begin{minipage}{\linewidth}
  \centering
  \includegraphics[width=\linewidth]{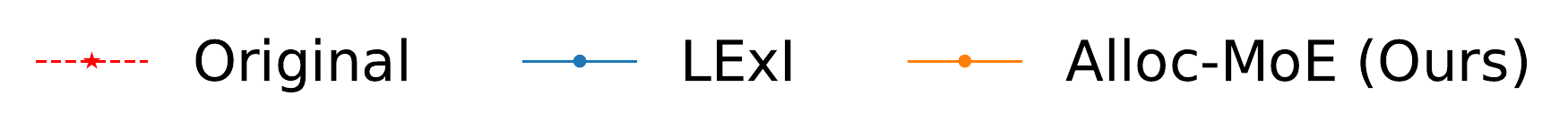}
\end{minipage}

\vspace{-0mm}

% ---------- (a) ----------
\begin{minipage}{0.48\linewidth}
  \centering
  \includegraphics[width=\linewidth]{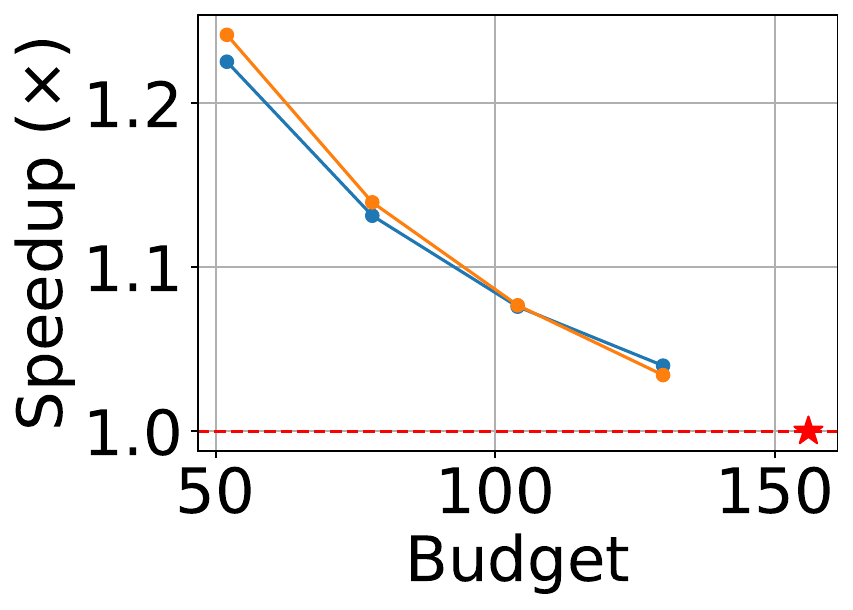}
  \vspace{-2mm}
  \small {\hspace{2.5em}(a)}
\end{minipage}\hfill
% ---------- (b) ----------
\begin{minipage}{0.48\linewidth}
  \centering
  \includegraphics[width=\linewidth]{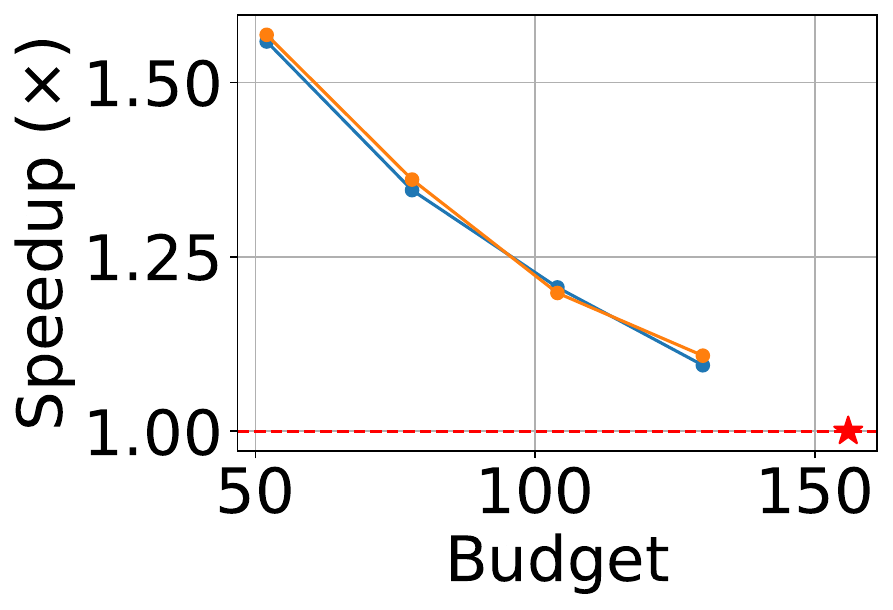}
  \vspace{-2mm}
  \small {\hspace{3em}(b)}
\end{minipage}\hfill

\caption{Speedup ratios for (a) prefill and (b) decode stages under varying global activation budgets.}
\label{fig:speedup}
\end{figure}

As shown in Figure~\ref{fig:speedup}, Alloc-MoE achieves inference speedups comparable to \emph{LExI} baseline across all global activation budgets in both prefill and decode stages, indicating that it introduces no additional runtime overhead.
Moreover, inference latency decreases monotonically as the budget is reduced, suggesting that the observed speedups mainly stem from reduced expert activations.
Under a representative setting where the budget is halved, Alloc-MoE achieves a $1.15\times$ speedup in prefill and a $1.34\times$ speedup in decode relative to the original expert activation allocation strategy, demonstrating consistent inference acceleration under constrained budgets.

\begin{figure*}[t]
\centering

% ---------- legend ----------
\begin{minipage}{\linewidth}
  \centering
  \includegraphics[width=0.7\linewidth]{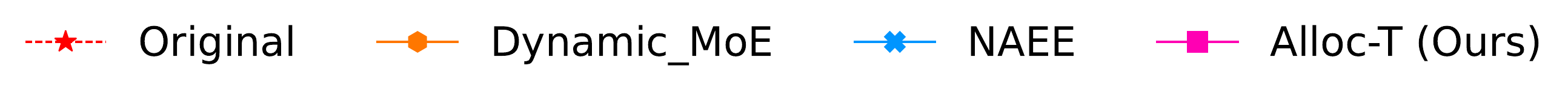}
\end{minipage}

\vspace{2mm}

% ---------- (a) ----------
\begin{minipage}{0.32\linewidth}
  \centering
  \includegraphics[width=\linewidth]{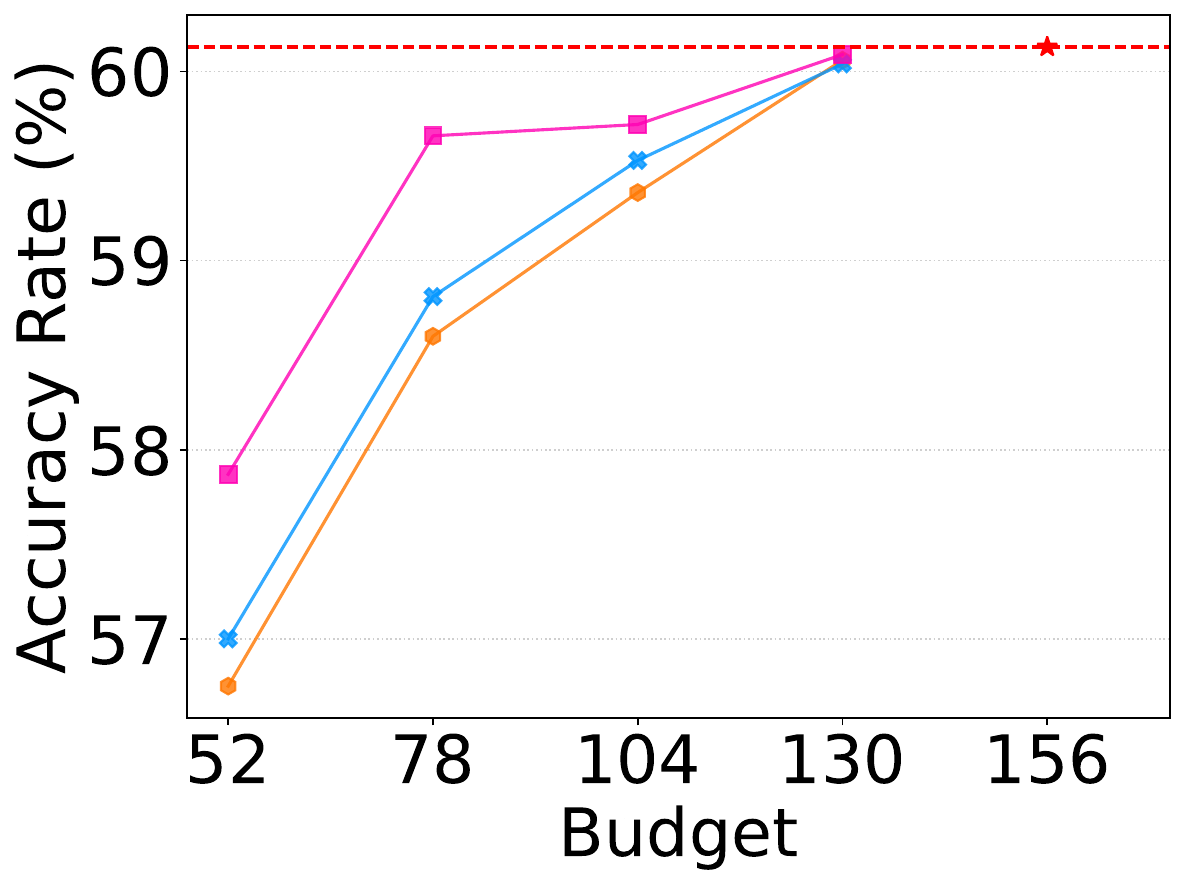}
  \vspace{-2mm}
  \small {\hspace{2em}(a)}
\end{minipage}\hfill
% ---------- (b) ----------
\begin{minipage}{0.32\linewidth}
  \centering
  \includegraphics[width=\linewidth]{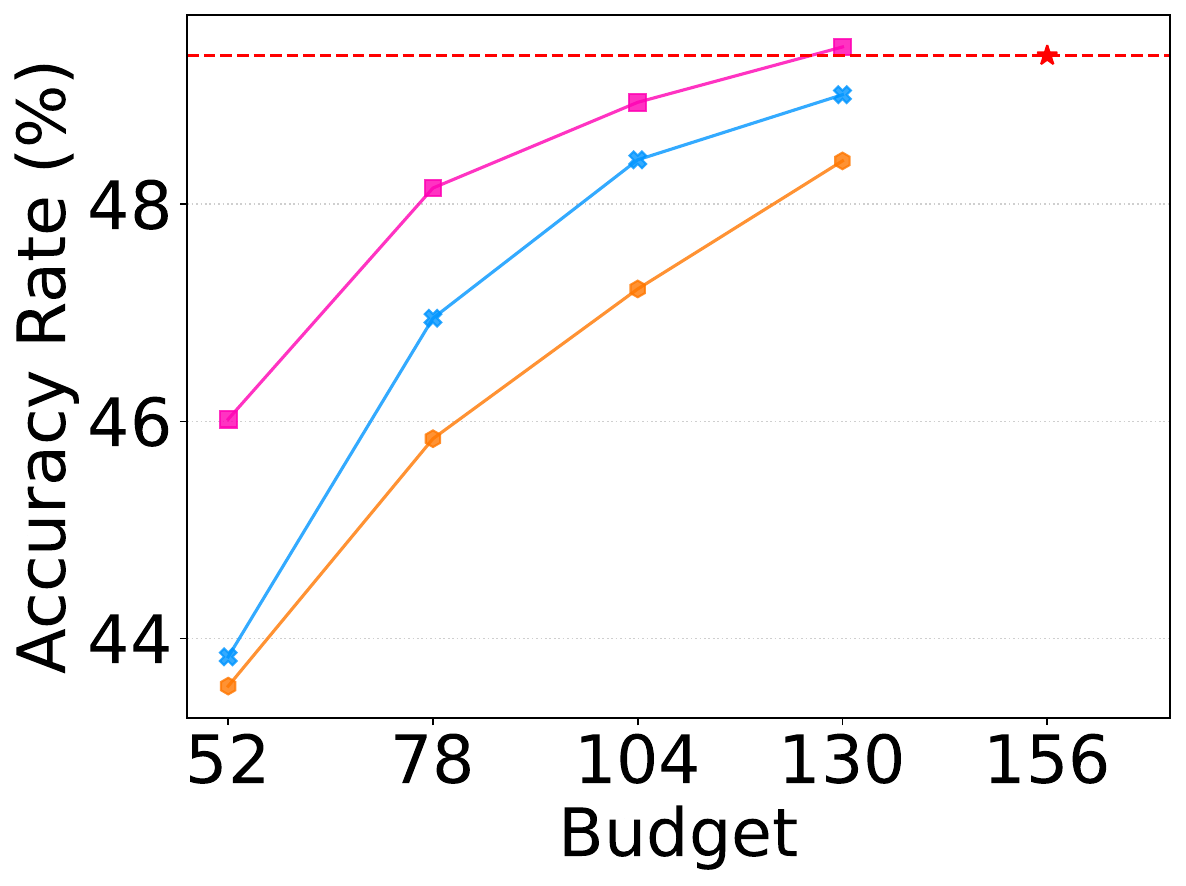}
  \vspace{-2mm}
  \small {\hspace{2em}(b)}
\end{minipage}\hfill
% ---------- (c) ----------
\begin{minipage}{0.32\linewidth}
  \centering
  \includegraphics[width=\linewidth]{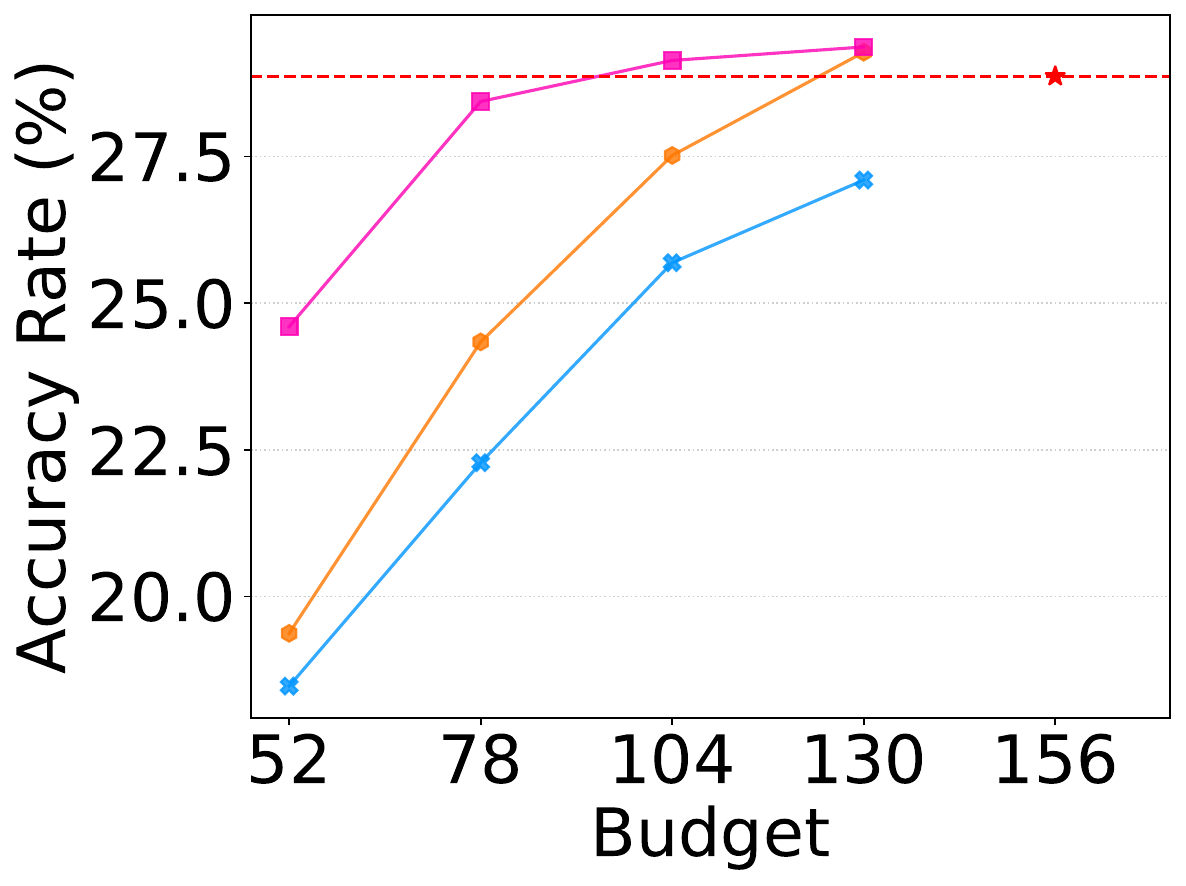}
  \vspace{-2mm}
  \small {\hspace{2em}(c)}
\end{minipage}
\vspace{-2mm}
\caption{Ablation results of Alloc-T on DeepSeek-V2-Lite under varying global activation budgets. (a) shows NLU task, (b) Reasoning task and (c) Math task.}
\label{fig:DSV2 Alloc-T}
\end{figure*}

\subsection{Analysis of the Base Expert Allocation $K_{base}$ in Alloc-T}

\begin{table}[t]
\centering
\resizebox{0.9\linewidth}{!}{
\begin{tabular}{lccccc}
\toprule
\multirow{2}{*}{\textbf{$K_{\text{base}}$}} 
% & \multicolumn{4}{c}{\textbf{Budget Ratio}}
& \multicolumn{4}{c}{\textbf{Budget}}
& \multirow{2}{*}{\textbf{Avg.}} \\
\cmidrule(lr){2-5}
& \textbf{52} & \textbf{78} & \textbf{104} & \textbf{130} &  \\
% & \textbf{2/6} & \textbf{3/6} & \textbf{4/6} & \textbf{5/6} &  \\
\midrule
0 & \underline{42.68} & \underline{45.36} & \textbf{45.99} & 46.09 & \underline{45.03} \\
\rowcolor{gray!15}
1 & \textbf{42.83} & \textbf{45.42} & \underline{45.93} & \textbf{46.30} & \textbf{45.12} \\
2 & 41.13 & 45.02 & 45.83 & \underline{46.17} & 44.54 \\
3 & 41.13 & 43.77 & 45.84 & 46.08 & 44.21 \\
4 & 41.13 & 43.77 & 45.52 & 45.80 & 44.06 \\
5 & 41.13 & 43.77 & 45.52 & 45.97 & 44.10 \\
% 6 & 41.13 & 43.77 & 45.52 & 45.97 & 44.10 \\
\bottomrule
\end{tabular}
}
\caption{Analysis of $K_{\text{base}}$ in Alloc-T on DeepSeek-V2-Lite under \emph{Uniform} allocation strategy.
Results are averaged across three task groups, with the last column showing the average performance across all budgets.}
\label{tab:kbase_ablation_dsv2}
\end{table}

Based on \emph{Uniform} allocation, we vary $K_{\text{base}}$ from 0 to $K-1$ where $K$ denotes the original Top-K of each model. 
Notably, $K_{base} = 0$ corresponds to no base expert activation allocation, where all activated experts are selected solely through expert activation redistribution.
To evaluate the overall impact of $K_{\text{base}}$, we report the average accuracy across three task groups.

Table \ref{tab:kbase_ablation_dsv2} summarizes the results on DeepSeek. 
Across most budget settings, $K_{\text{base}}=1$ achieves the best or near-best performance and produces the highest average accuracy across all four budgets. 
In contrast, larger base allocations ($K_{base} \geq 2$) consistently degrade performance, especially under tighter budgets, suggesting that excessive base allocation over-constrains the allocation space and limits the effectiveness of adaptive expert allocation.
While $K_{base}=0$ performs competitively, it remains slightly inferior to $K_{base}=1$, which indicates that $K_{\text{base}}=1$ consistently offers the most favorable trade-off between base allocation and flexible redistribution.
Similar trends are observed on Qwen and OLMoE in Appendix~\ref{subsec:additional_results} (Tables~\ref{tab:kbase_ablation_qwen} and~\ref{tab:kbase_ablation_olmoe}).
Based on these observations, we set $K_{\text{base}}=1$ as the default configuration of Alloc-T. 

\subsection{Analysis of Expert Load Balance}

\begin{figure}[t]
\centering
\includegraphics[width=\linewidth]{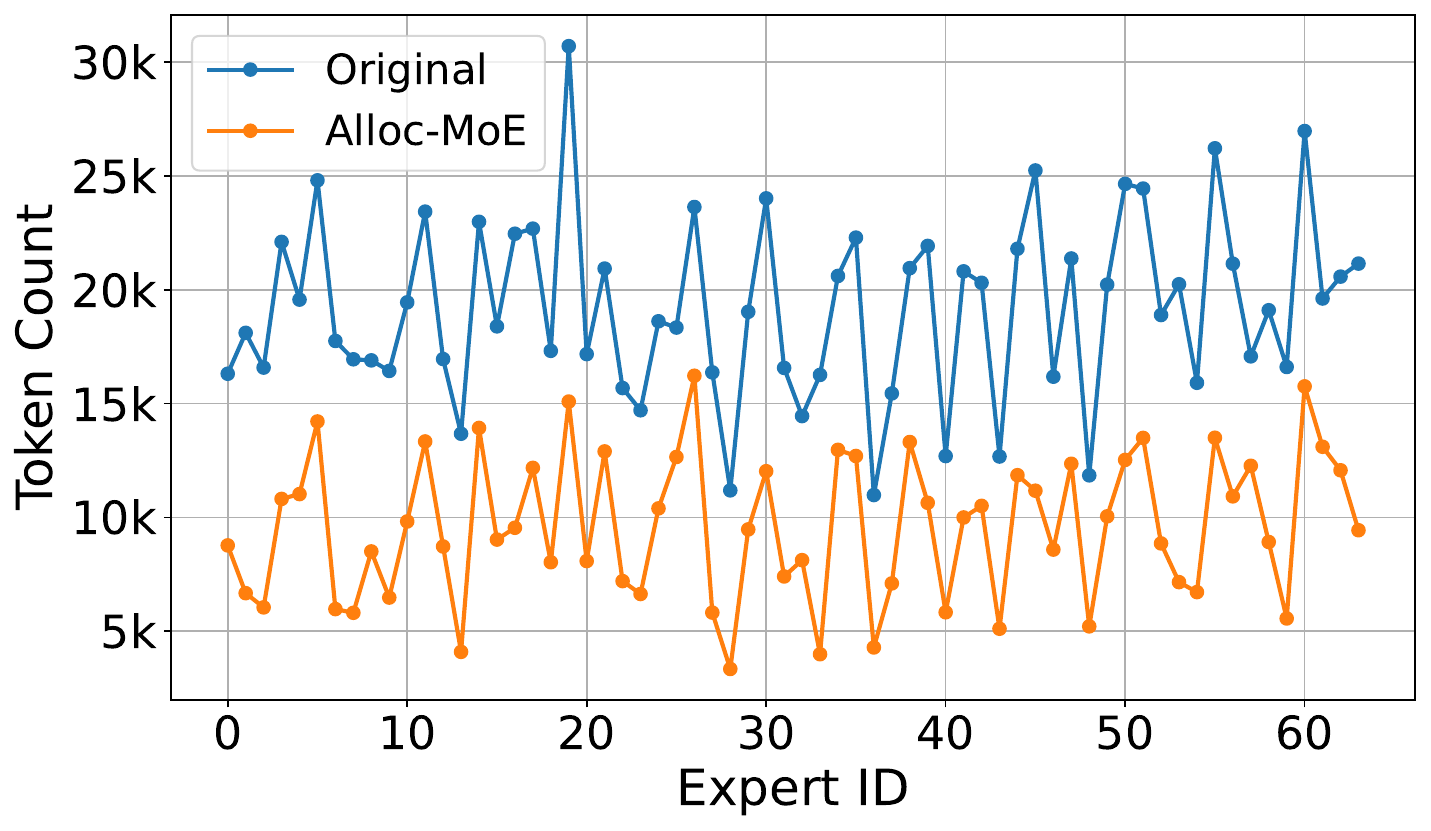}
\caption{Expert load distributions in the 10-th MoE layer of DeepSeek-V2-Lite under the original configuration and Alloc-MoE with Budget = 78. Alloc-MoE preserves the load pattern while reducing per-expert load, potentially lowering inter-device communication overhead.}
\label{fig:expert_load}
\end{figure}

\begin{figure}[t]
\centering
\includegraphics[width=\linewidth]{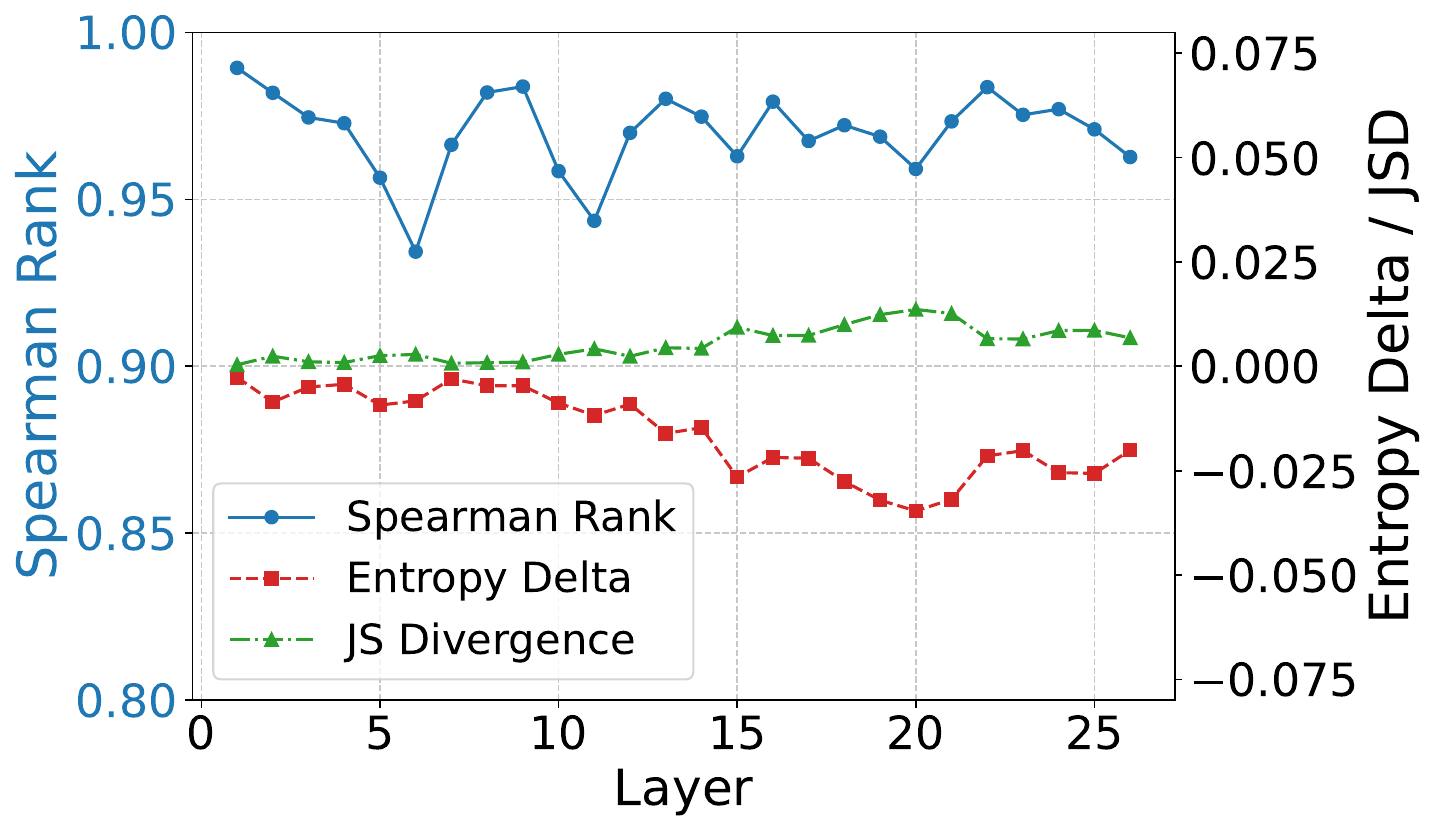}
\caption{Analysis of load balance across layers. Including Spearman rank correlation (blue, left axis), Entropy delta (red, right axis), and JS divergence (green, right axis).}
\label{fig:metrics_variation}
\end{figure}

Figure~\ref{fig:expert_load} illustrates the result of expert load distributions. Qualitatively, Alloc-MoE preserves the overall shape of the load distribution, introducing no noticeable distortion. 
Moreover, by reducing the per-expert load, it decreases the communication volume, which is expected to improve inference efficiency in distributed MoE deployment.
We further conduct a quantitative analysis to characterize this behavior, as summarized in Figure~\ref{fig:metrics_variation}. 
Specifically, we compute the \emph{Spearman rank correlation} of expert loads across all layers between the two settings. The correlation remains consistently high (0.93--0.99), indicating that the relative ordering of \emph{hot} and \emph{cold} experts is largely preserved. This suggests that Alloc-MoE does not disrupt the inherent expert specialization.
To further assess distributional shifts, we measure the difference in normalized entropy between the two settings, as well as the Jensen–Shannon (JS) divergence between the corresponding weighted load distributions. Both metrics show minimal deviation: the entropy decreases slightly (0.003--0.035), while the JS divergence remains below 0.014 across all layers. 
These results demonstrate that Alloc-MoE maintains a stable expert utilization pattern while introducing negligible distributional shift.

\subsection{Analysis of Budget Allocation in Alloc-L and Alloc-T}

Under the half-budget setting, As shown in Appendix~\ref{subsec:additional_results} (Figure~\ref{fig:allocation_across_layers_and_tokens}), Alloc-L produces a clearly non-uniform layer-wise allocation, where earlier layers retain more experts while deeper layers operate with reduced budgets, reflecting heterogeneous routing sensitivity across layers. 
At the token level, Alloc-T shows strong correlation ($>0.7$) with routing entropy and exhibits a consistent monotonic pattern, allocating fewer experts to low-entropy (high-confidence) tokens and more to high-entropy (ambiguous) ones.

\subsection{Analysis of Calibration Dataset}
We compare the impact of different calibration datasets, including WikiText2~\cite{wikitext}, C4~\cite{raffel2020c4}, and Pile~\cite{gao2020pile}. As shown in Appendix~\ref{subsec:additional_results} (Table~\ref{tab:calib_dataset}), the overall performance remains highly consistent across all three datasets under varying budget settings, indicating that Alloc-L is insensitive to the selection of calibration datasets.

\begin{table}[t]
\centering
\resizebox{0.95\linewidth}{!}{
\begin{tabular}{lccccc}
\toprule
\multirow{2}{*}{\textbf{Dataset}} 
& \multicolumn{4}{c}{\textbf{Budget}}
& \multirow{2}{*}{\textbf{Avg.}} \\
\cmidrule(lr){2-5}
& \textbf{52} & \textbf{78} & \textbf{104} & \textbf{130} &  \\
\midrule
\textbf{WikiText2} & 41.53 & 44.61 & 45.76 & 46.47 & 44.59 \\
\textbf{C4}        & 41.56 & 45.05 & 45.83 & 45.94 & 44.60 \\
\textbf{Pile}      & 42.55 & 45.05 & 45.29 & 45.84 & 44.68 \\
\bottomrule
\end{tabular}
}
\caption{Impact of different calibration datasets on DeepSeek-V2-Lite with Alloc-L under varying budgets.}
\label{tab:calib_dataset}
\end{table}

\subsection{Ablation Studies}

We first evaluate Alloc-L and Alloc-T independently against their respective level-specific baselines to demonstrate their individual effectiveness.
We then analyze their joint behavior under the \emph{Uniform} allocations to highlight their complementarity.

\begin{table}[t]
\centering
\resizebox{0.95\linewidth}{!}{
\begin{tabular}{lccccc}
\toprule
\multirow{2}{*}{\textbf{Method}} 
& \multicolumn{4}{c}{\textbf{Budget}}
& \multirow{2}{*}{\textbf{Avg.}} \\
\cmidrule(lr){2-5}
& \textbf{52} & \textbf{78} & \textbf{104} & \textbf{130} &  \\
\midrule
\textbf{Uniform} & 41.27 & 43.90 & 45.51 & 46.06 & 44.19 \\
\textbf{+L} & 41.53 & 44.61 & 45.76 & \textbf{46.47} & 44.59 \\
\textbf{+T} & \underline{42.83} & \underline{45.42} & \underline{45.93} & \underline{46.30} & \underline{45.12} \\
\rowcolor{gray!15}
\textbf{+L +T} & \textbf{43.09} & \textbf{45.48} & \textbf{46.01} & 46.17 & \textbf{45.19} \\
\bottomrule
\end{tabular}
}
\caption{Ablation study of Alloc-L (\textbf{+L}) and Alloc-T (\textbf{+T}) on DeepSeek-V2-Lite under varying global activation budgets. 
Results are averaged across three task groups, with the last column showing the average performance across all budgets.}
\label{tab:ablation_dsv2}
\end{table}

\paragraph{Effect of Alloc-L.}
We compare Alloc-L against four layer-wise allocation baselines under four budget settings.
In addition to \emph{LExI} and \emph{Uniform}, we consider:
\emph{Ascending}, where the allocations increase with layer depth, and \emph{Descending}, which applies the reverse schedule.
Figure~\ref{fig:DSV2 Alloc-L} presents the results on DeepSeek.
Compared to these baselines, Alloc-L consistently achieves a superior performance–efficiency trade-off.
% by more effectively distributing expert activation across layers under fixed budgets.
Similar trends are observed on Qwen and OLMoE in Appendix~\ref{subsec:additional_results} (Figure~\ref{fig:Qwen and OLMoE Alloc-L}).
Across both models, Alloc-L outperforms the baseline strategies in the majority of budget configurations, indicating that Alloc-L generalizes well across different MoE models.

\paragraph{Effect of Alloc-T.}
We compare Alloc-T against \emph{NAEE} and \emph{Dynamic-MoE} under the \emph{Uniform} layer-wise allocations. Figure~\ref{fig:DSV2 Alloc-T} presents the results on DeepSeek.
Alloc-T consistently outperforms both baselines across all tasks and budgets.
Notably, its advantage becomes increasingly pronounced under tighter budgets and on more challenging tasks, highlighting the benefit of token-wise redistribution under aggressive sparsification.
Similar trends are observed on Qwen and OLMoE as shown in Appendix~\ref{subsec:additional_results} (Figure~\ref{fig:Qwen and OLMoE Alloc-T}), indicating that Alloc-T generalizes well across different MoE models.

\paragraph{Complementarity of Alloc-L and Alloc-T.}
Table~\ref{tab:ablation_dsv2} reports the ablation results on DeepSeek.
The results show that: 
(1) Alloc-L consistently improves performance across all budget settings, achieving an average gain of 0.4\%.
% by allocating expert activations in a layer-aware manner under constrained activation budgets.
(2) Alloc-T yields larger improvements than Alloc-L in most budget settings, particularly under aggressive sparsification, highlighting the growing importance of token-level redistribution as the budget becomes increasingly constrained.
(3) Combining Alloc-L and Alloc-T consistently achieves the best or near-best performance across all budgets, with the highest average performance.
Similar trends are observed on Qwen and OLMoE in Appendix~\ref{subsec:additional_results} (Tables~\ref{tab:ablation_qwen} and ~\ref{tab:ablation_olmoe}), indicating that the benefits of Alloc-L and Alloc-T generalize across different MoE models.
Overall, these results suggest that Alloc-L and Alloc-T operate on orthogonal dimensions of expert activation allocation and can be jointly applied to improve the allocation of limited budgets.

\section{Conclusion}
In this work, we proposed Alloc-MoE, a unified framework that optimizes the allocation of limited expert activation in Mixture-of-Experts models to minimize performance degradation.
By modeling expert activations as a global activation budget and allocating it in a coordinated manner across layers and tokens, Alloc-MoE effectively mitigates performance degradation caused by reduced expert activations. Extensive experiments across multiple MoE models and tasks demonstrate that Alloc-MoE consistently achieves a superior performance--efficiency trade-off, preserving accuracy even with substantially fewer activated experts.

\section{Limitations}
While Alloc-MoE demonstrates strong performance under constrained expert activation budgets, several limitations remain. 
First, while our approach focuses on allocating expert activations, it is fully orthogonal to other efficiency-oriented methods such as expert pruning or quantization, which could be combined with Alloc-MoE for additional speedups. 
Second, Alloc-MoE does not incorporate hardware-level factors such as expert placement or communication overhead. Integrating these considerations in distributed systems represents a natural direction for future enhancement. 
Finally, our framework targets pretrained models, and extending Alloc-MoE to incorporate activation-aware strategies during training to improve model robustness remains an open avenue for further research.

% \section*{Acknowledgments}

\bibliography{paper}

\appendix
\section{Setup Details}
\subsection{Implementation Details of Baselines}
\label{subsec:heuristic_allocation}

\paragraph{Layer-level Baseline.}
The \emph{Ascending} baseline allocates a total budget $B$ across $L$ layers following an ascending allocation schedule. It first initializes an integer allocation vector via rounded linear interpolation between $K_{\min}$ and $K_{\max}$, yielding a depth-increasing allocation profile. The allocation is then refined through bi-directional passes to enforce monotonicity and a maximum inter-layer step size of one. To exactly satisfy the global activation budget constraint $B$, we apply greedy adjustments to internal layers, incrementing or decrementing allocations only when monotonicity and boundary constraints are preserved.
The \emph{Descending} allocation is obtained by applying the same procedure in reverse order.
In most settings, we set $K_{\min}=1$ and $K_{\max}$ to the original Top-K of the model, allowing the budget to be allocated primarily through depth-dependent variation. Only when the global budget $B$ exceeds what can be accommodated under these bounds do we increase $K_{\min}$ accordingly, until a feasible allocation satisfying the budget constraint is obtained.

\paragraph{Token-level Baseline.}
\emph{Dynamic-MoE} performs dynamic token routing by adjusting the number of expert activations per token based on a pre-profiled Top-$P$ threshold. The threshold is calibrated under the \emph{Uniform} layer-level allocation, ensuring that the resulting expert activations meet the target budget.
\emph{NAEE} implements Dynamic Expert Skipping by conditionally skipping experts based on routing scores. In our evaluation, we extend its original Top-2 routing to Top-K routing, skipping the $k$-th expert if its score falls below a layer- and model-specific threshold $\beta$ relative to the Top-1 expert. This threshold is profiled under the \emph{Uniform} layer-wise allocation to ensure the same target expert budget is maintained.

\subsection{Efficiency Evaluation Details}
\label{subsec:eval_details}
For inference efficiency evaluation, we measure prefill and decode speedup ratios using dummy prompts with randomly generated tokens (batch size 8, prompt length 32, decode length 128).
This configuration provides a controlled and reproducible environment that isolates the impact of expert allocation on inference latency. Speedups are averaged over 10 runs after a 5 warm-up iterations to reduce variability, and reported relative to the original configuration.

\section{Additional Results}
\label{subsec:additional_results}
\begin{table}[h]
\centering
\resizebox{0.9\linewidth}{!}{
\begin{tabular}{lcccccc}
\toprule
\multirow{2}{*}{\textbf{$K_{\text{base}}$}}
& \multicolumn{4}{c}{\textbf{Budget}} 
& \multirow{2}{*}{\textbf{Avg.}} \\
\cmidrule(lr){2-5}
& \textbf{48} & \textbf{60} & \textbf{72} & \textbf{84} &  \\
\midrule
0 & \underline{46.87} & 47.86 & 48.06 & 48.63 & \underline{47.86} \\
\rowcolor{gray!15}
1 & \textbf{46.89} & 47.84 & 48.23 & \underline{48.66} & \textbf{47.91} \\
2 & 46.00 & \textbf{47.93} & \textbf{48.33} & \textbf{48.69} & 47.74 \\
3 & 46.00 & \underline{47.90} & \underline{48.25} & 48.48 & 47.66 \\
% 4 & 46.00 & \underline{47.90} & \underline{48.25} & 48.53 & 47.67 \\
\bottomrule
\end{tabular}
}
\caption{Analysis of $K_{\text{base}}$ in Alloc-T on Qwen1.5-MoE-A2.7B under \emph{Uniform} allocation strategy.
Results are averaged across three task groups, with the last column showing the average performance across all budgets.}
\label{tab:kbase_ablation_qwen}
\end{table}

\begin{table}[h]
\centering
\resizebox{0.9\linewidth}{!}{
\begin{tabular}{lcccccc}
\toprule
\multirow{2}{*}{\textbf{$K_{\text{base}}$}} 
& \multicolumn{4}{c}{\textbf{Budget}} 
& \multirow{2}{*}{\textbf{Avg.}} \\
\cmidrule(lr){2-5}
& \textbf{64} & \textbf{80} & \textbf{96} & \textbf{112} &  \\
\midrule
0 & 37.27 & 38.47 & \underline{39.14} & \underline{39.55} & 38.61 \\
\rowcolor{gray!15}
1 & 37.38 & \underline{38.50} & 39.09 & \textbf{39.61} & \underline{38.65} \\
2 & \underline{37.39} & 38.49 & \textbf{39.16} & \underline{39.55} & \underline{38.65} \\
3 & \textbf{37.53} & \textbf{38.56} & 39.05 & 39.51 & \textbf{38.66} \\
4 & 36.77 & \textbf{38.56} & 39.04 & 39.49 & 38.47 \\
5 & 36.77 & 37.97 & 38.96 & 39.61 & 38.33 \\
6 & 36.77 & 37.97 & 39.11 & 39.54 & 38.35 \\
7 & 36.77 & 37.97 & 39.11 & 39.47 & 38.33 \\
% 8 & 36.77 & 37.97 & 39.11 & 39.47 & 38.33 \\
\bottomrule
\end{tabular}
}
\caption{Analysis of the $K_{\text{base}}$ in Alloc-T on OLMoE-1B-7B-0924 under \emph{Uniform} allocation strategy.
Results are averaged across three task groups, with the last column showing the average performance across all budgets.}
\label{tab:kbase_ablation_olmoe}
\end{table}

\begin{table}[h]
\centering
\setlength{\tabcolsep}{4.5pt}  % 默认约6pt
\resizebox{0.9\linewidth}{!}{
\begin{tabular}{lccccc}
\toprule
\multirow{2}{*}{\textbf{Method}} 
& \multicolumn{4}{c}{\textbf{Budget}}
& \multirow{2}{*}{\textbf{Avg.}} \\
\cmidrule(lr){2-5}
& \textbf{48} & \textbf{60} & \textbf{72} & \textbf{84} &  \\
\midrule
\textbf{Uniform} & 46.07 & 47.35 & 48.18 & 48.10 & 47.43 \\
\textbf{+L} & 46.11 & \underline{47.78} & \textbf{48.33} & 48.59 & 47.70 \\
\textbf{+T} & \underline{46.89} & \textbf{47.84} & \underline{48.19} & \textbf{48.66} & \textbf{47.90} \\
\rowcolor{gray!15}
\textbf{+L +T} & \textbf{46.96} & 47.54 & \underline{48.19} & \underline{48.63} & \underline{47.83} \\
\bottomrule
\end{tabular}
}
\caption{Ablation study of Alloc-L (\textbf{+L}) and Alloc-T (\textbf{+T}) on Qwen1.5-MoE-A2.7B under varying global activation budgets. 
Results are averaged across three task groups, with the last column showing the average performance across all budgets.}
\label{tab:ablation_qwen}
\end{table}

\begin{table}[h]
\centering
\setlength{\tabcolsep}{4.5pt}  % 默认约6pt
\resizebox{0.9\linewidth}{!}{
\begin{tabular}{lccccc}
\toprule
\multirow{2}{*}{\textbf{Method}} 
& \multicolumn{4}{c}{\textbf{Budget}}
& \multirow{2}{*}{\textbf{Avg.}} \\
\cmidrule(lr){2-5}
& \textbf{64} & \textbf{80} & \textbf{96} & \textbf{112} &  \\
\midrule
\textbf{Uniform} & 36.70 & 37.83 & 38.93 & 39.30 & 38.19 \\
\textbf{+L} & 37.08 & 38.21 & \underline{39.09} & 39.35 & 38.43 \\
\textbf{+T} & \underline{37.38} & \underline{38.50} & 39.06 & \textbf{39.61} & \underline{38.64} \\
\rowcolor{gray!15}
\textbf{+L +T} & \textbf{37.53} & \textbf{38.59} & \textbf{39.27} & \underline{39.53} & \textbf{38.73} \\
\bottomrule
\end{tabular}
}
\caption{Ablation study of Alloc-L (\textbf{+L}) and Alloc-T (\textbf{+T}) on OLMoE-1B-7B-0924 under varying global activation budgets. 
Results are averaged across three task groups, with the last column showing the average performance across all budgets.}
\label{tab:ablation_olmoe}
\end{table}

\begin{figure*}[t]
\centering

% ---------- legend ----------
\begin{minipage}{\linewidth}
  \centering
  \includegraphics[width=0.9\linewidth]{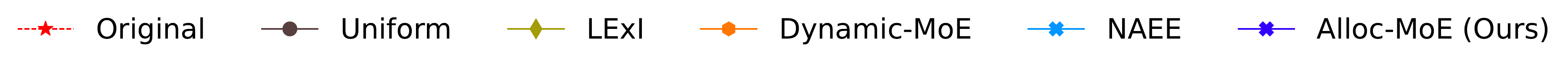}
\end{minipage}

\vspace{2mm}

% ---------- (a) ----------
\begin{minipage}{0.32\linewidth}
  \centering
  \includegraphics[width=\linewidth]{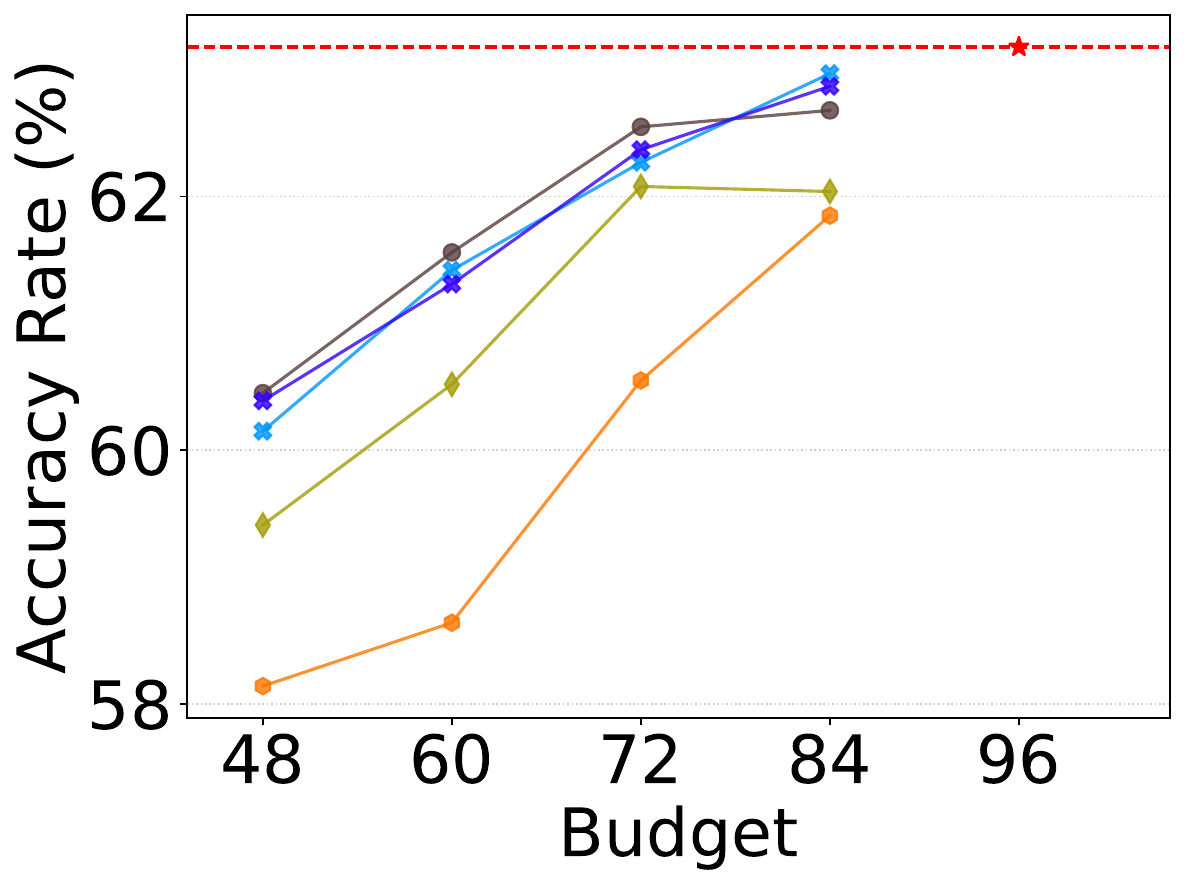}
  \vspace{-2mm}
  \small {\hspace{2em}(a)}
\end{minipage}\hfill
% ---------- (b) ----------
\begin{minipage}{0.32\linewidth}
  \centering
  \includegraphics[width=\linewidth]{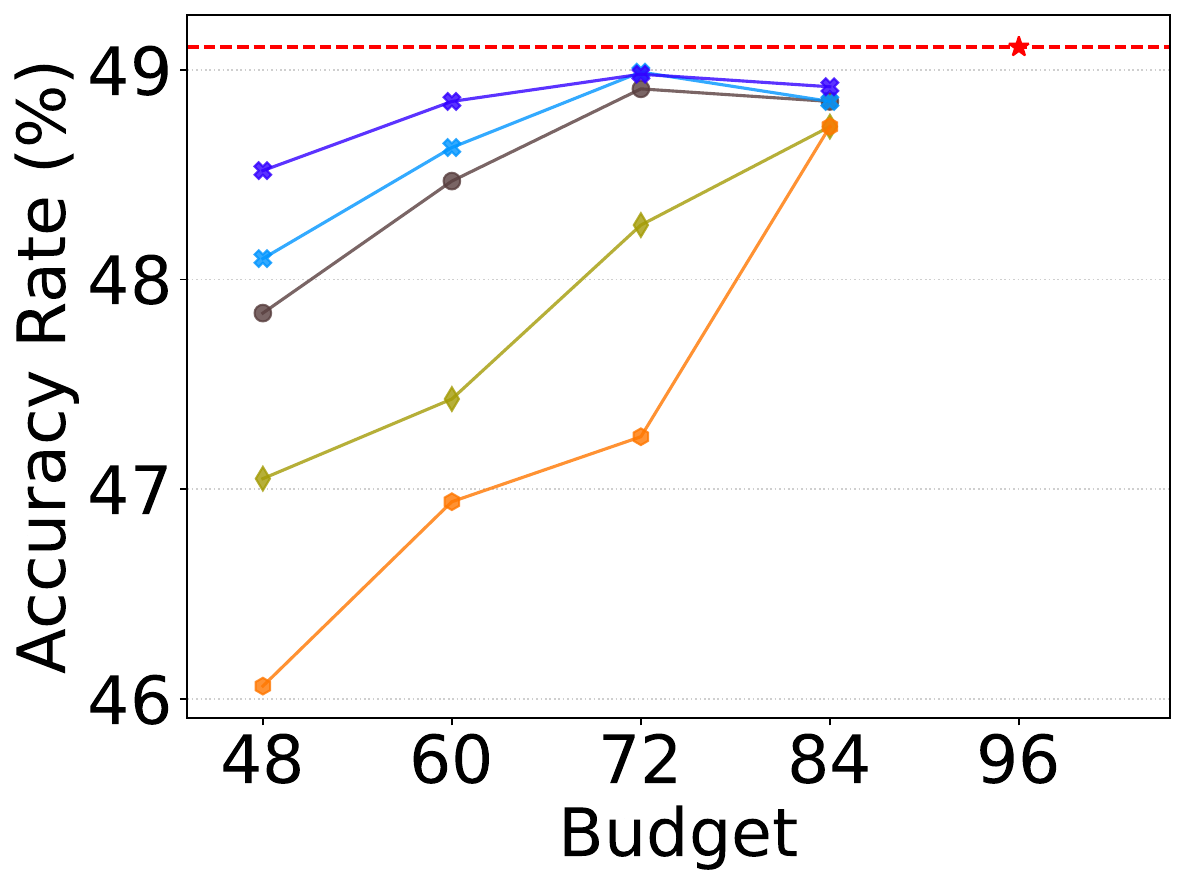}
  \vspace{-2mm}
  \small {\hspace{2em}(b)}
\end{minipage}\hfill
% ---------- (c) ----------
\begin{minipage}{0.32\linewidth}
  \centering
  \includegraphics[width=\linewidth]{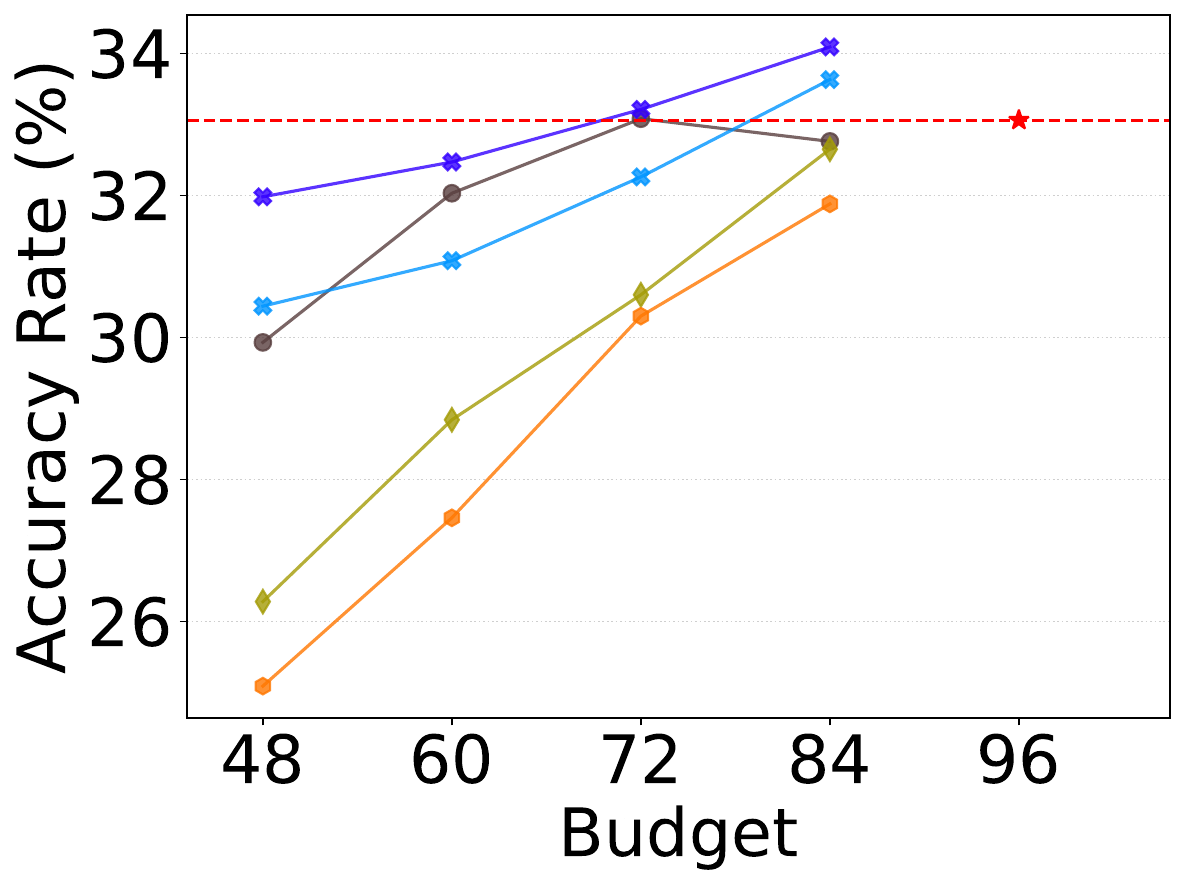}
  \vspace{-2mm}
  \small {\hspace{2em}(c)}
\end{minipage}

\vspace{2mm}

\begin{minipage}{0.32\linewidth}
  \centering
  \includegraphics[width=\linewidth]{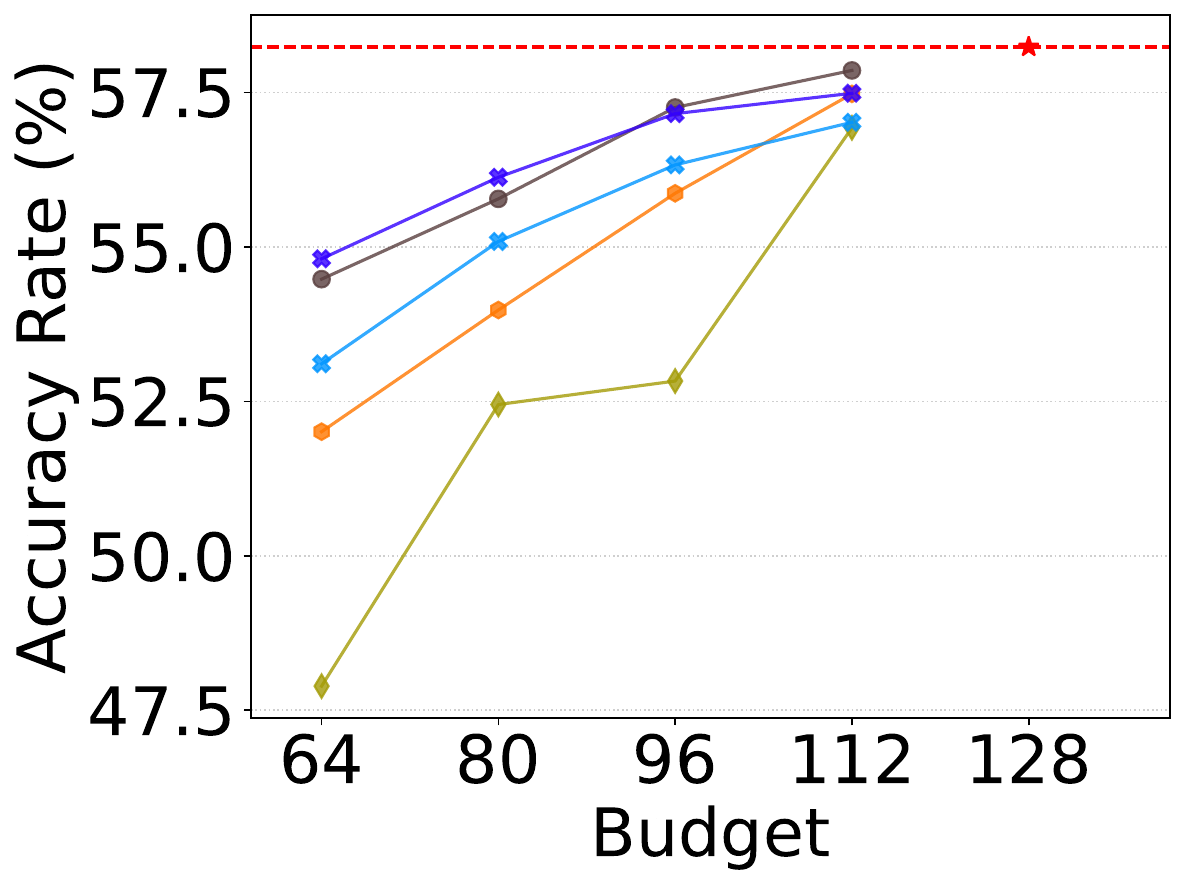}
  \vspace{-2mm}
  \small {\hspace{3em}(d)}
\end{minipage}\hfill
% ---------- (b) ----------
\begin{minipage}{0.32\linewidth}
  \centering
  \includegraphics[width=\linewidth]{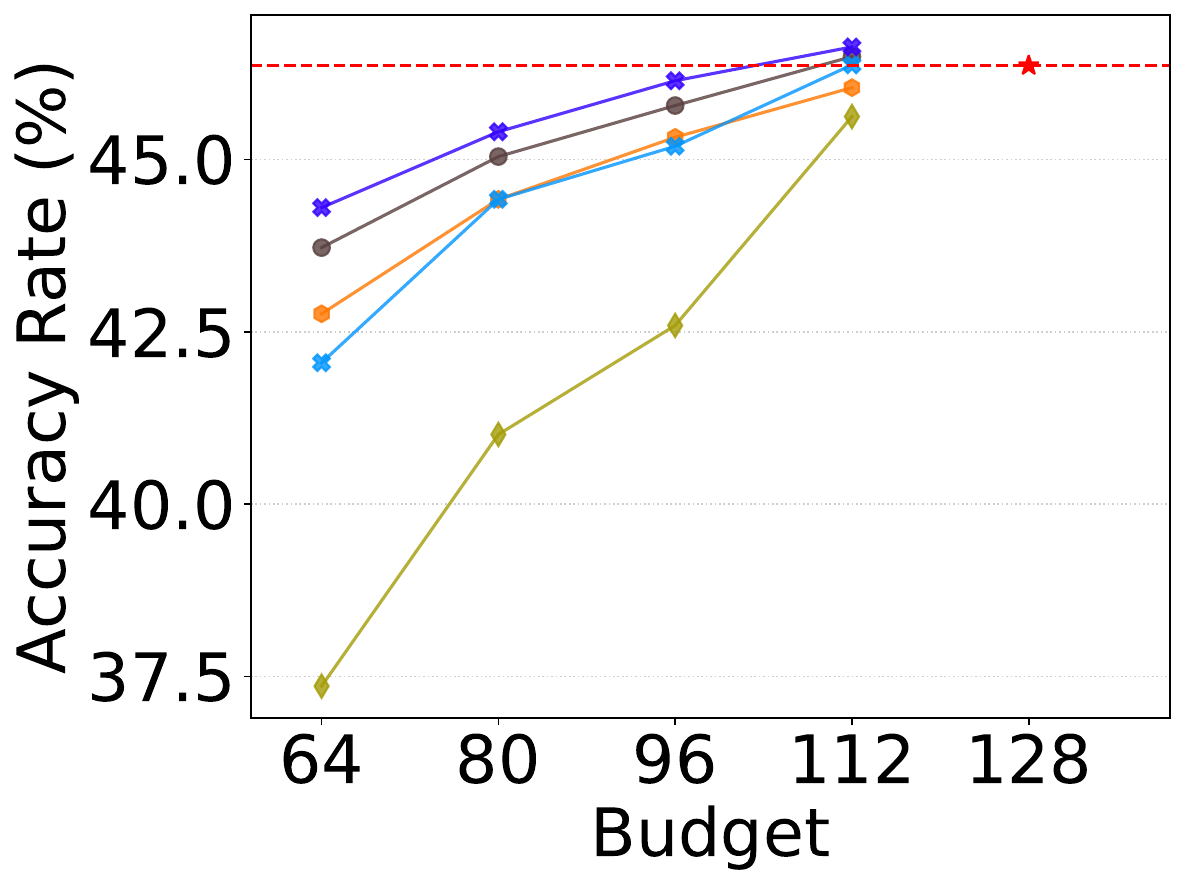}
  \vspace{-2mm}
  \small {\hspace{2em}(e)}
\end{minipage}\hfill
% ---------- (c) ----------
\begin{minipage}{0.32\linewidth}
  \centering
  \includegraphics[width=\linewidth]{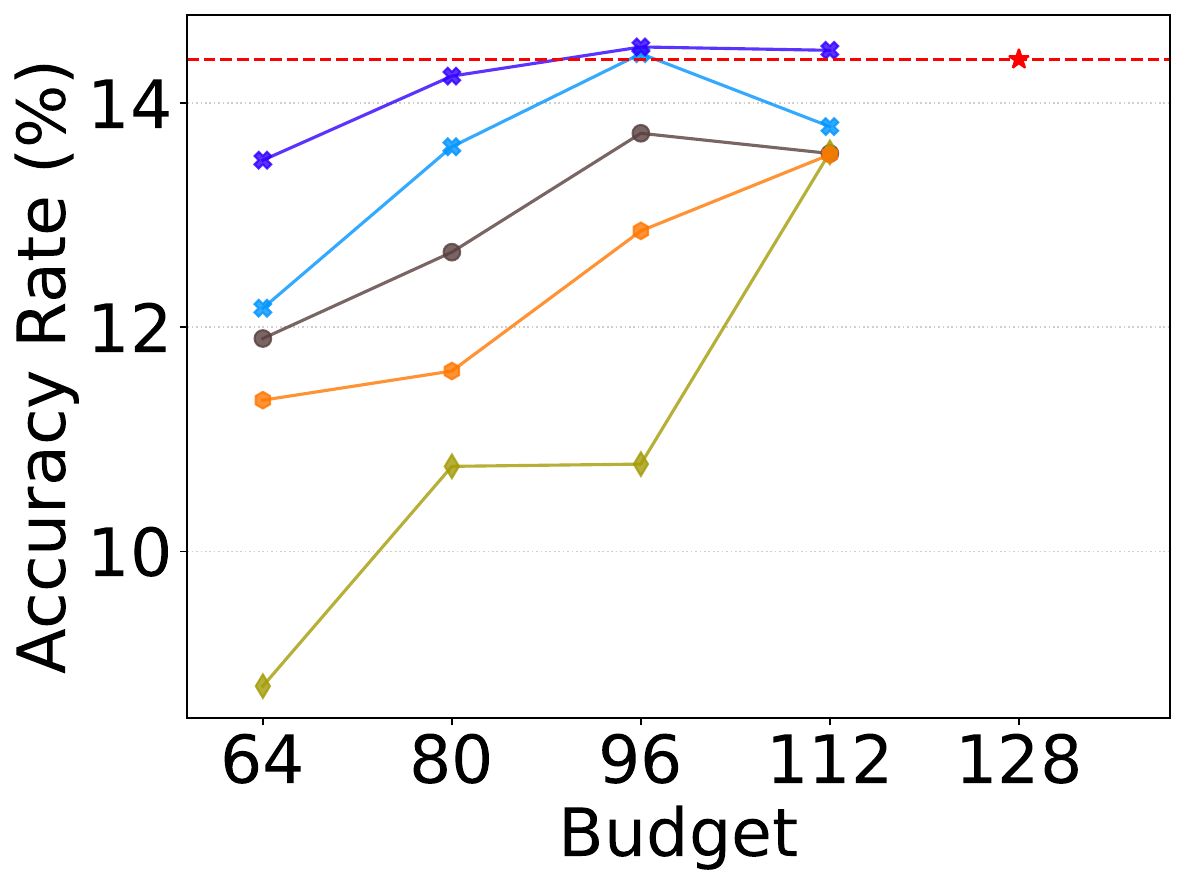}
  \vspace{-2mm}
  \small {\hspace{2em}(f)}
\end{minipage}

\caption{Alloc-MoE results on Qwen1.5-MoE-A2.7B and OLMoE-1B-7B-0924 under varying global activation budgets.
(a–c) report results on Qwen1.5-MoE-A2.7B, while (d–f) correspond to OLMoE-1B-7B-0924.
For each model, (a,d) show NLU tasks, (b,e) Reasoning tasks, and (c,f) Math tasks.}
\label{fig:Qwen and OLMoE Alloc-MoE}
\end{figure*}

\begin{figure*}[t]
\centering

% ---------- legend ----------
\begin{minipage}{\linewidth}
  \centering
  \includegraphics[width=0.9\linewidth]{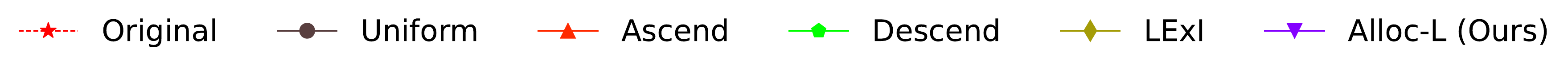}
\end{minipage}

\vspace{2mm}

% ---------- (a) ----------
\begin{minipage}{0.32\linewidth}
  \centering
  \includegraphics[width=\linewidth]{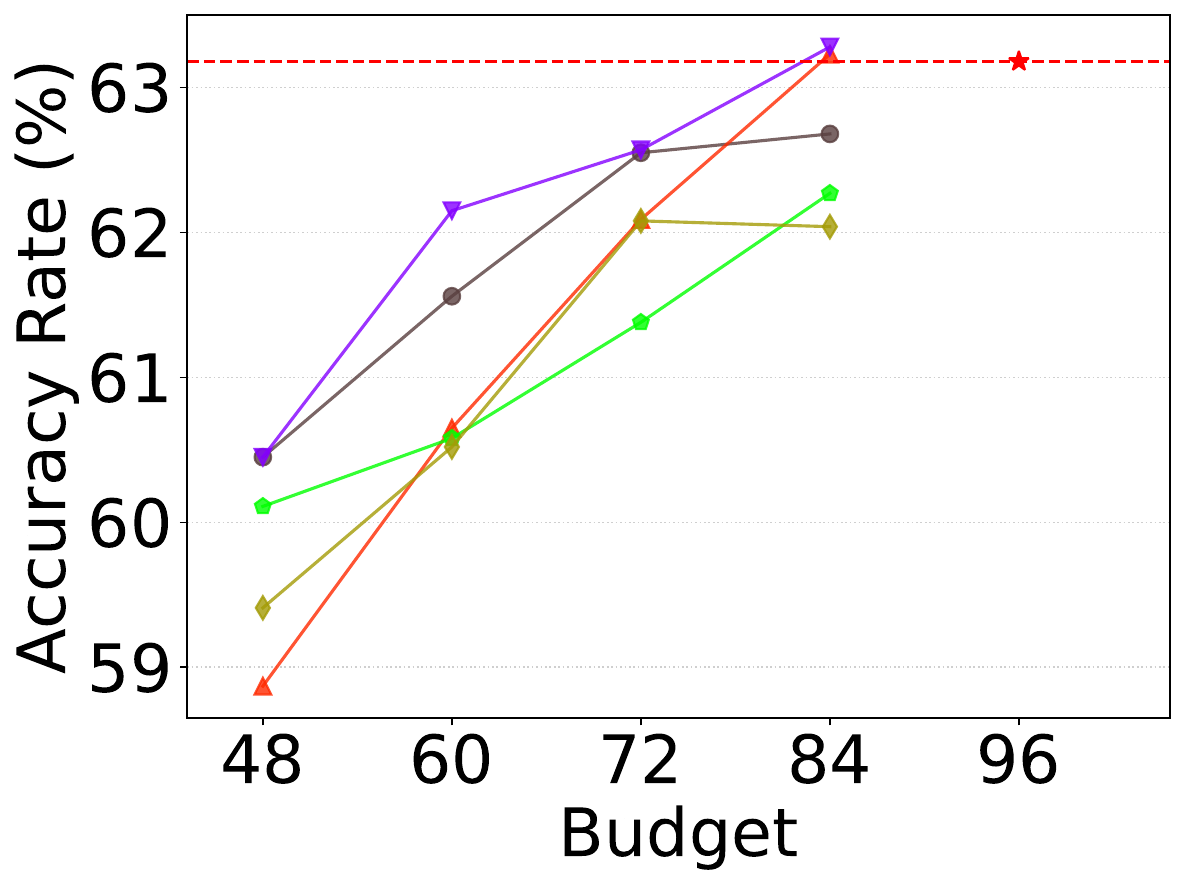}
  \vspace{-2mm}
  \small {\hspace{2em}(a)}
\end{minipage}\hfill
% ---------- (b) ----------
\begin{minipage}{0.32\linewidth}
  \centering
  \includegraphics[width=\linewidth]{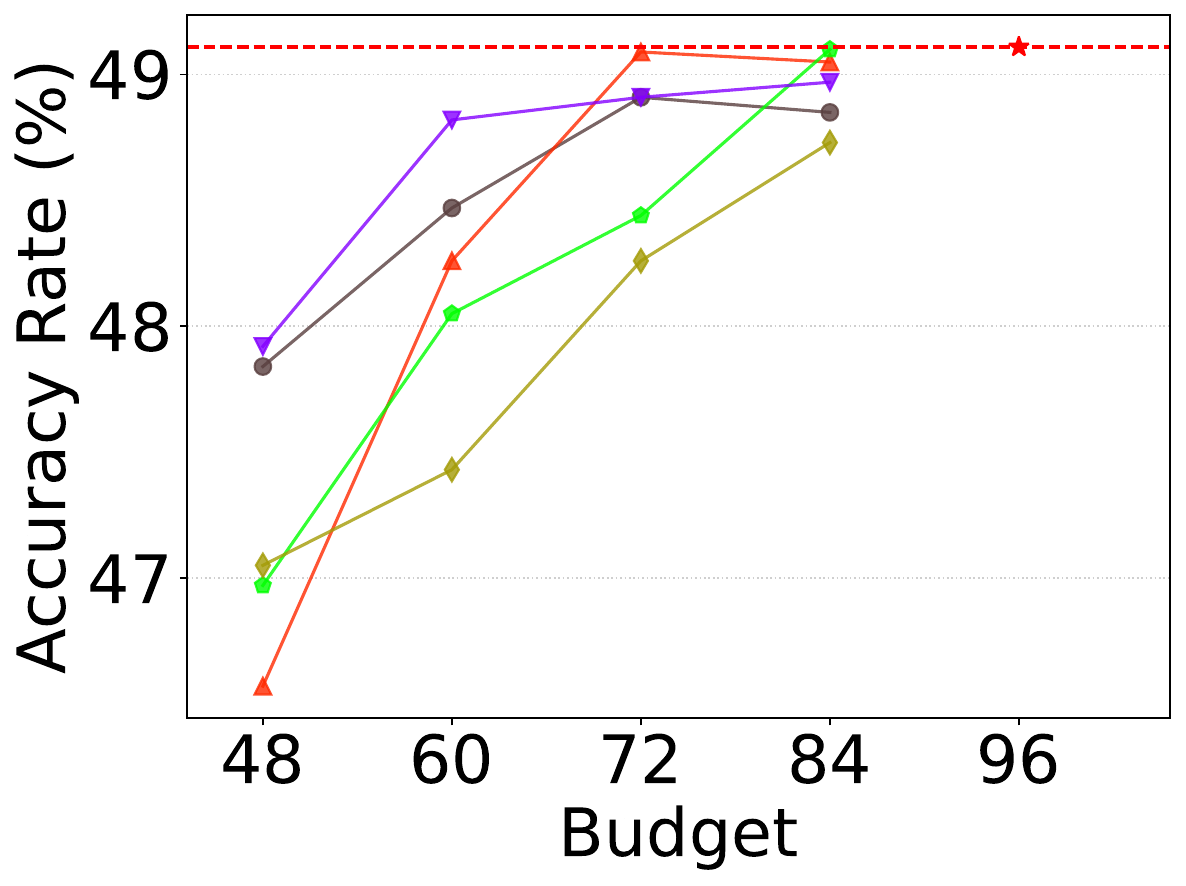}
  \vspace{-2mm}
  \small {\hspace{2em}(b)}
\end{minipage}\hfill
% ---------- (c) ----------
\begin{minipage}{0.32\linewidth}
  \centering
  \includegraphics[width=\linewidth]{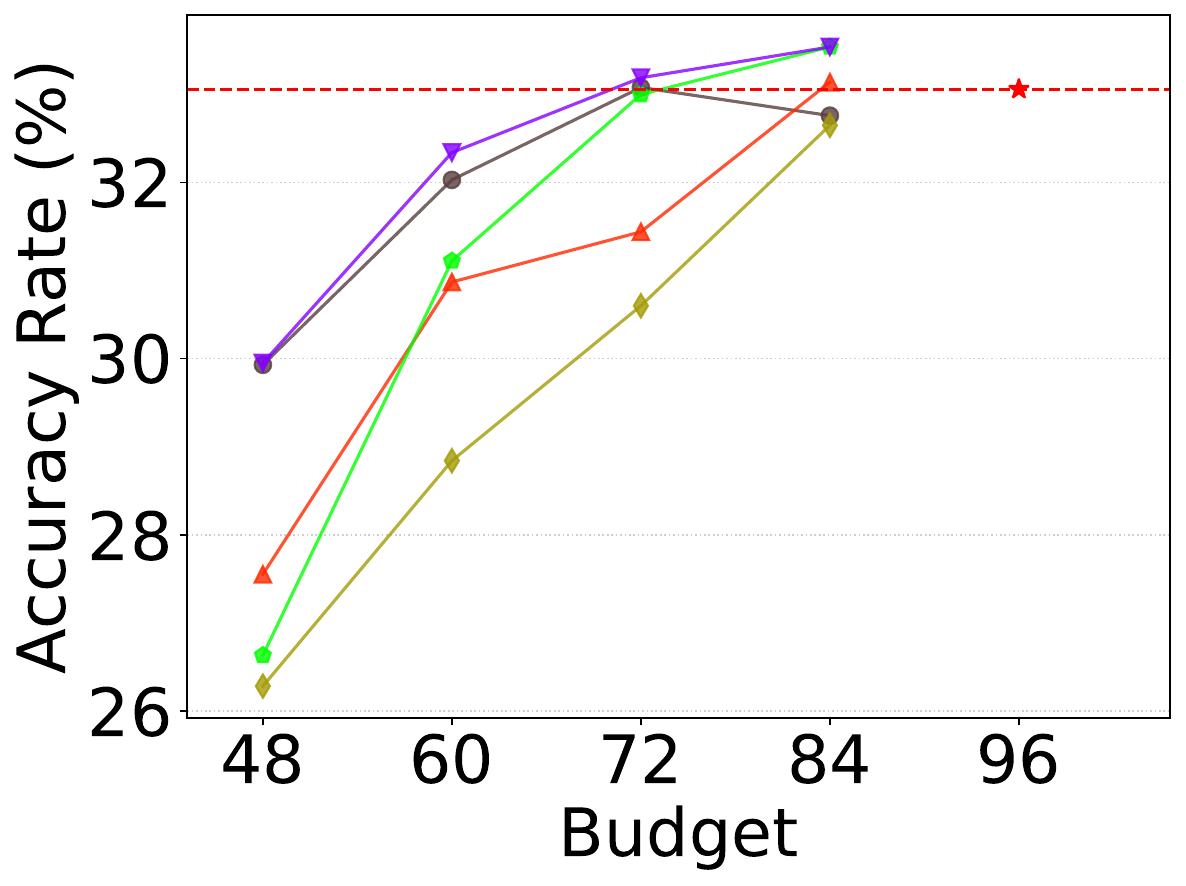}
  \vspace{-2mm}
  \small {\hspace{2em}(c)}
\end{minipage}

\vspace{2mm}

\begin{minipage}{0.32\linewidth}
  \centering
  \includegraphics[width=\linewidth]{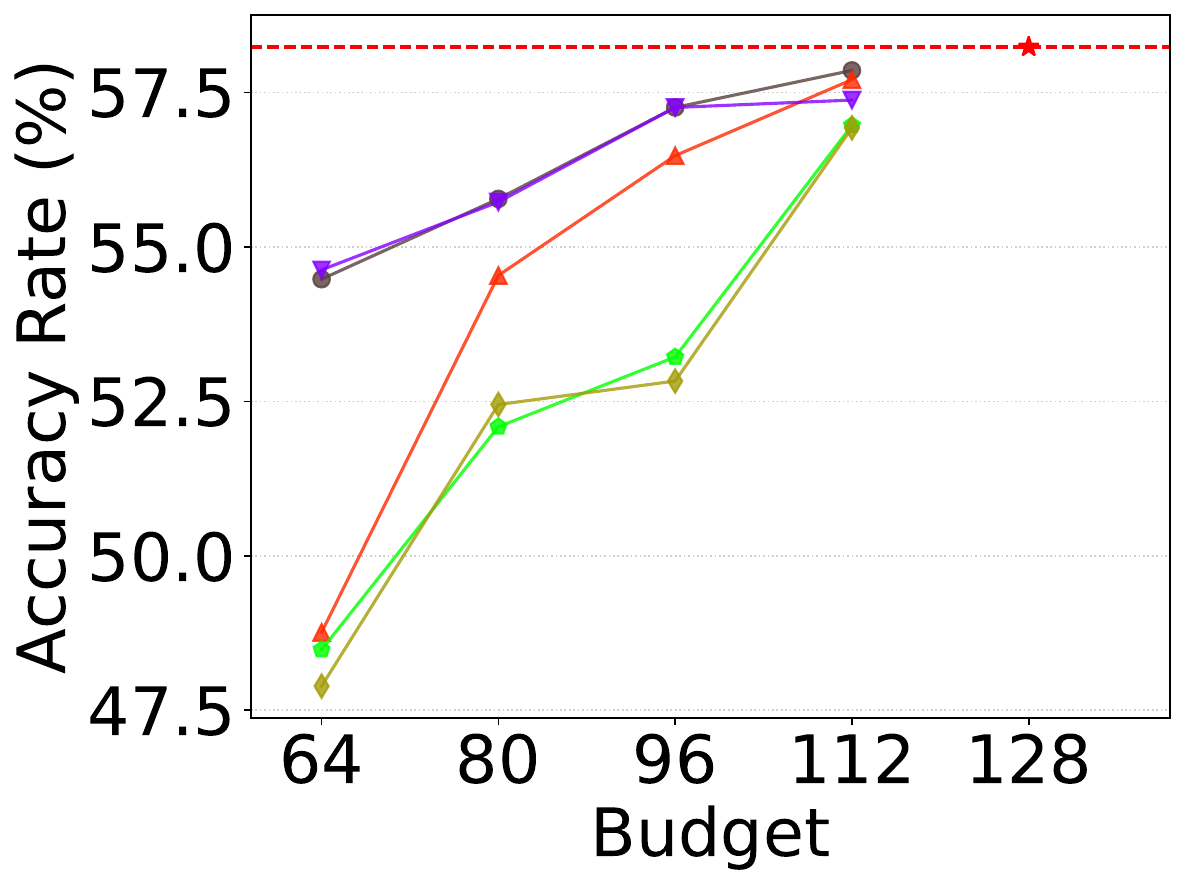}
  \vspace{-2mm}
  \small {\hspace{3em}(d)}
\end{minipage}\hfill
% ---------- (b) ----------
\begin{minipage}{0.32\linewidth}
  \centering
  \includegraphics[width=\linewidth]{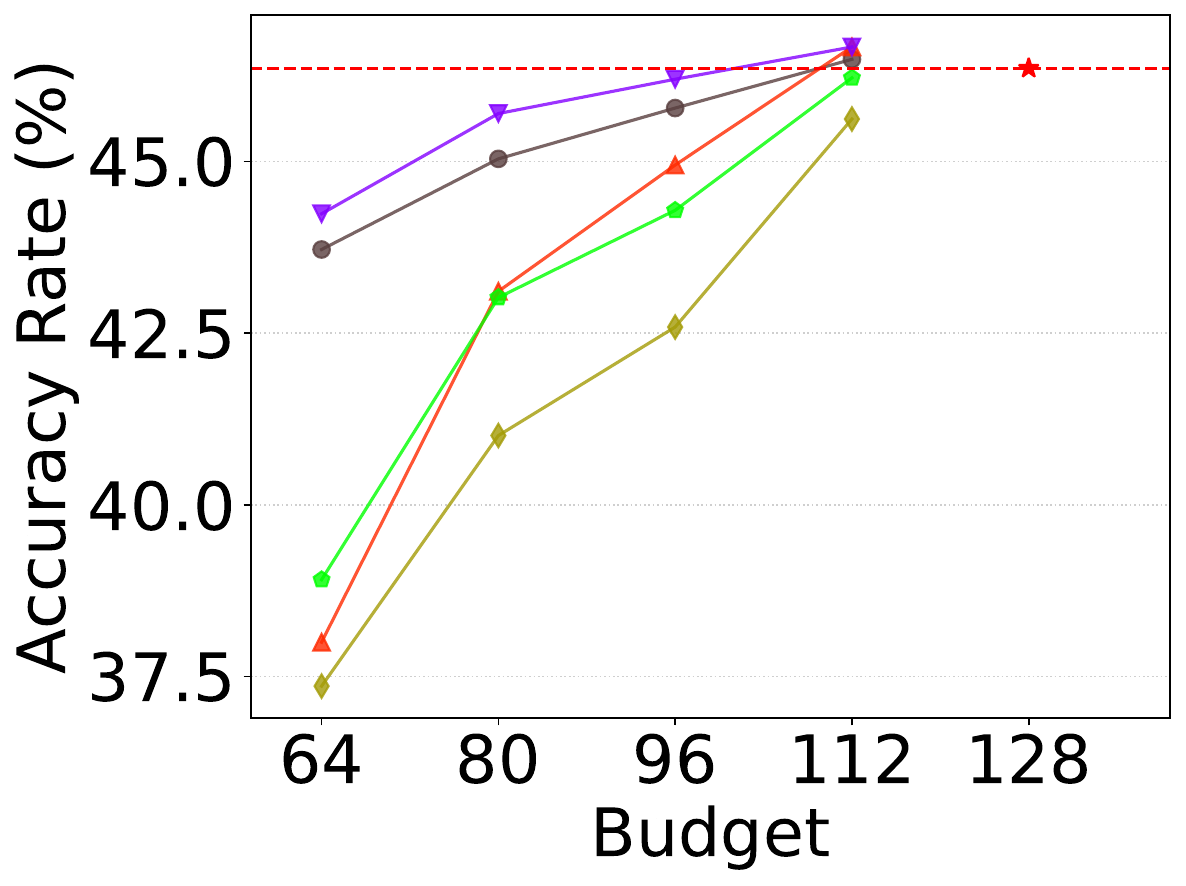}
  \vspace{-2mm}
  \small {\hspace{3em}(e)}
\end{minipage}\hfill
% ---------- (c) ----------
\begin{minipage}{0.32\linewidth}
  \centering
  \includegraphics[width=\linewidth]{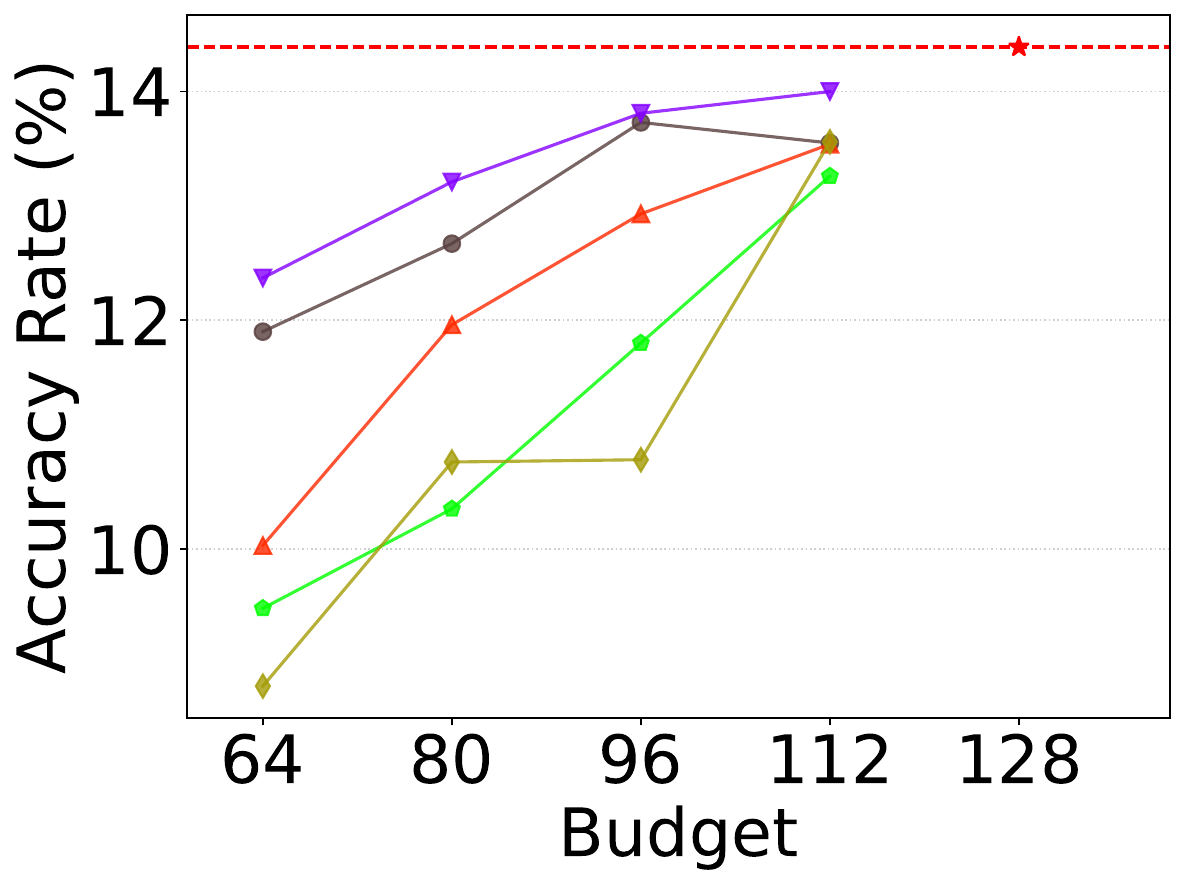}
  \vspace{-2mm}
  \small {\hspace{2em}(f)}
\end{minipage}
\caption{Ablation results of Alloc-L on Qwen1.5-MoE-A2.7B and OLMoE-1B-7B-0924 under varying global activation budgets.
(a–c) report results on Qwen1.5-MoE-A2.7B, while (d–f) correspond to OLMoE-1B-7B-0924.
For each model, (a,d) show NLU tasks, (b,e) Reasoning tasks, and (c,f) Math tasks.}
\label{fig:Qwen and OLMoE Alloc-L}
\end{figure*}

\begin{figure*}[t]
\centering

% ---------- legend ----------
\begin{minipage}{\linewidth}
  \centering
  \includegraphics[width=0.7\linewidth]{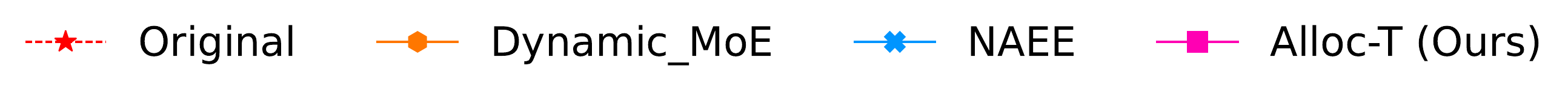}
\end{minipage}

\vspace{2mm}

% ---------- (a) ----------
\begin{minipage}{0.32\linewidth}
  \centering
  \includegraphics[width=\linewidth]{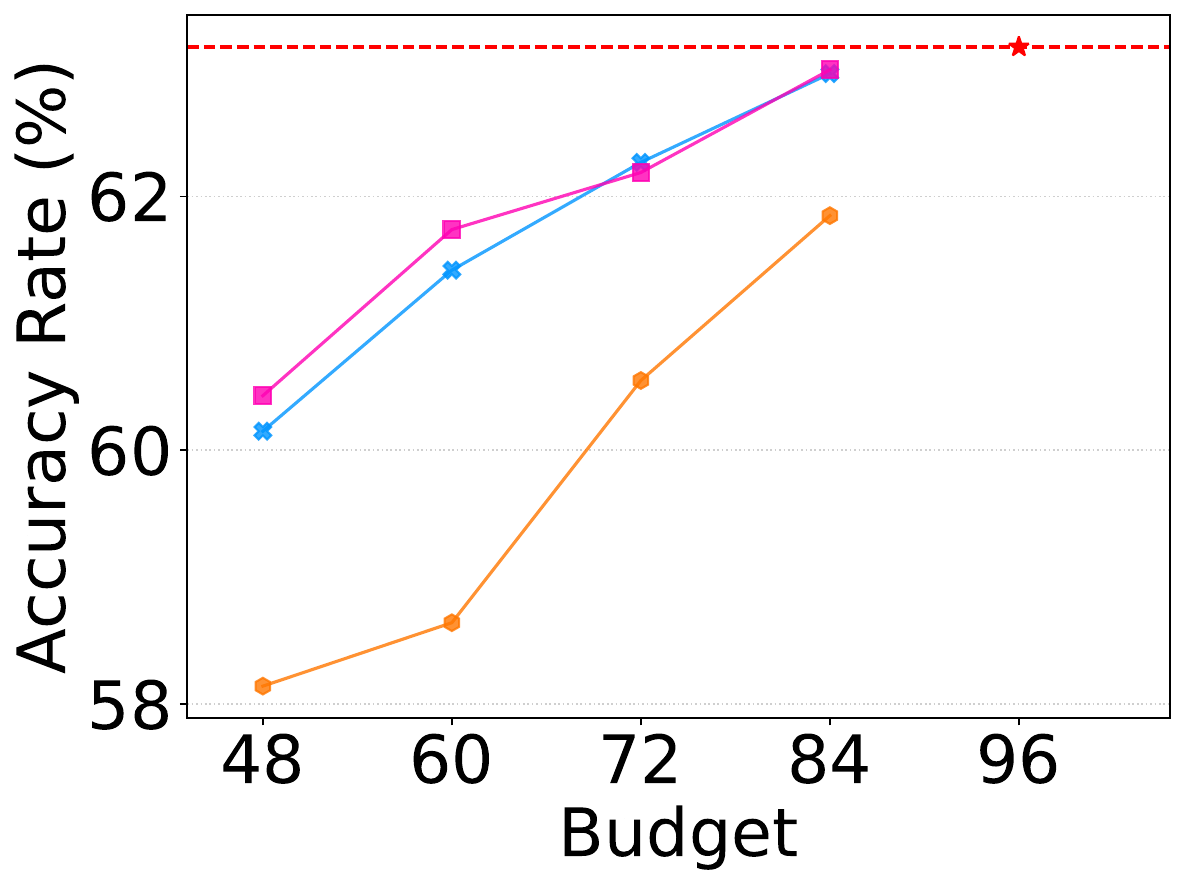}
  \vspace{-2mm}
  \small {\hspace{2em}(a)}
\end{minipage}\hfill
% ---------- (b) ----------
\begin{minipage}{0.32\linewidth}
  \centering
  \includegraphics[width=\linewidth]{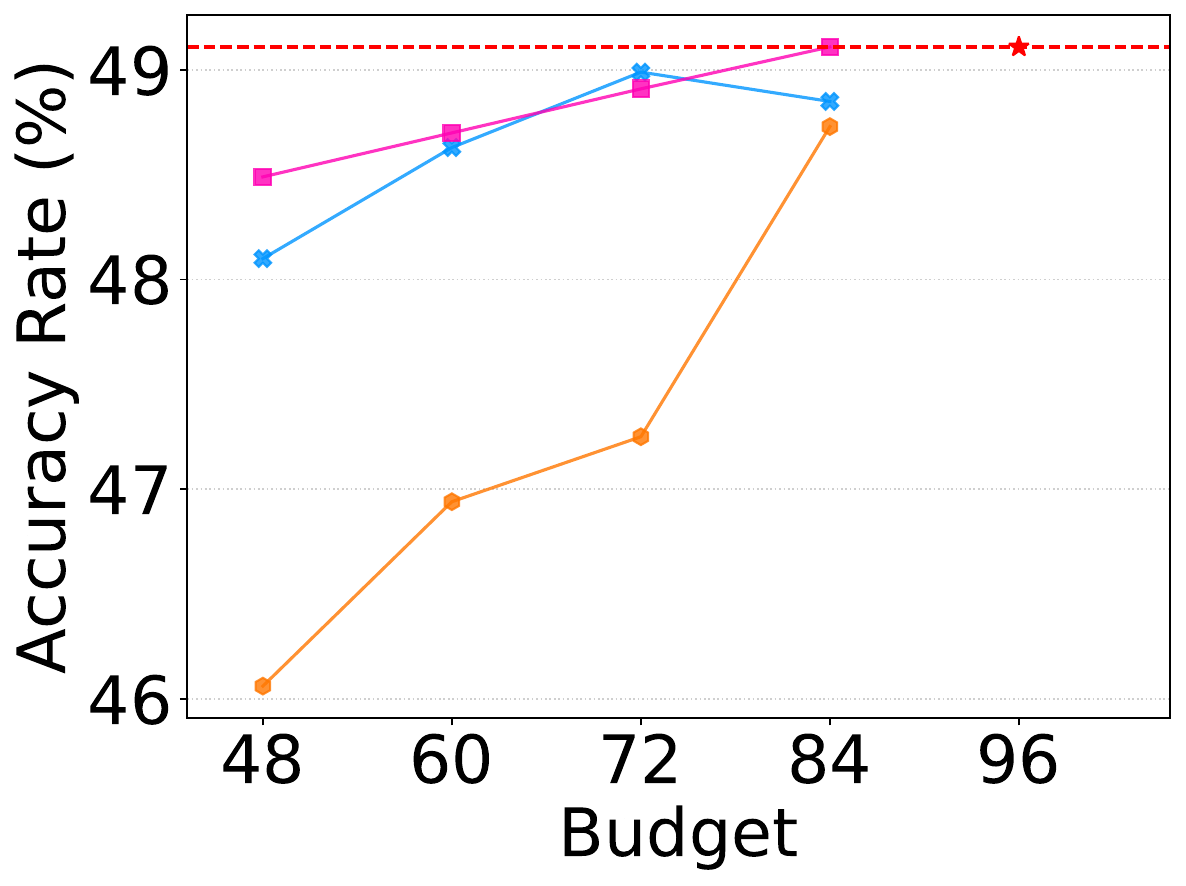}
  \vspace{-2mm}
  \small {\hspace{2em}(b)}
\end{minipage}\hfill
% ---------- (c) ----------
\begin{minipage}{0.32\linewidth}
  \centering
  \includegraphics[width=\linewidth]{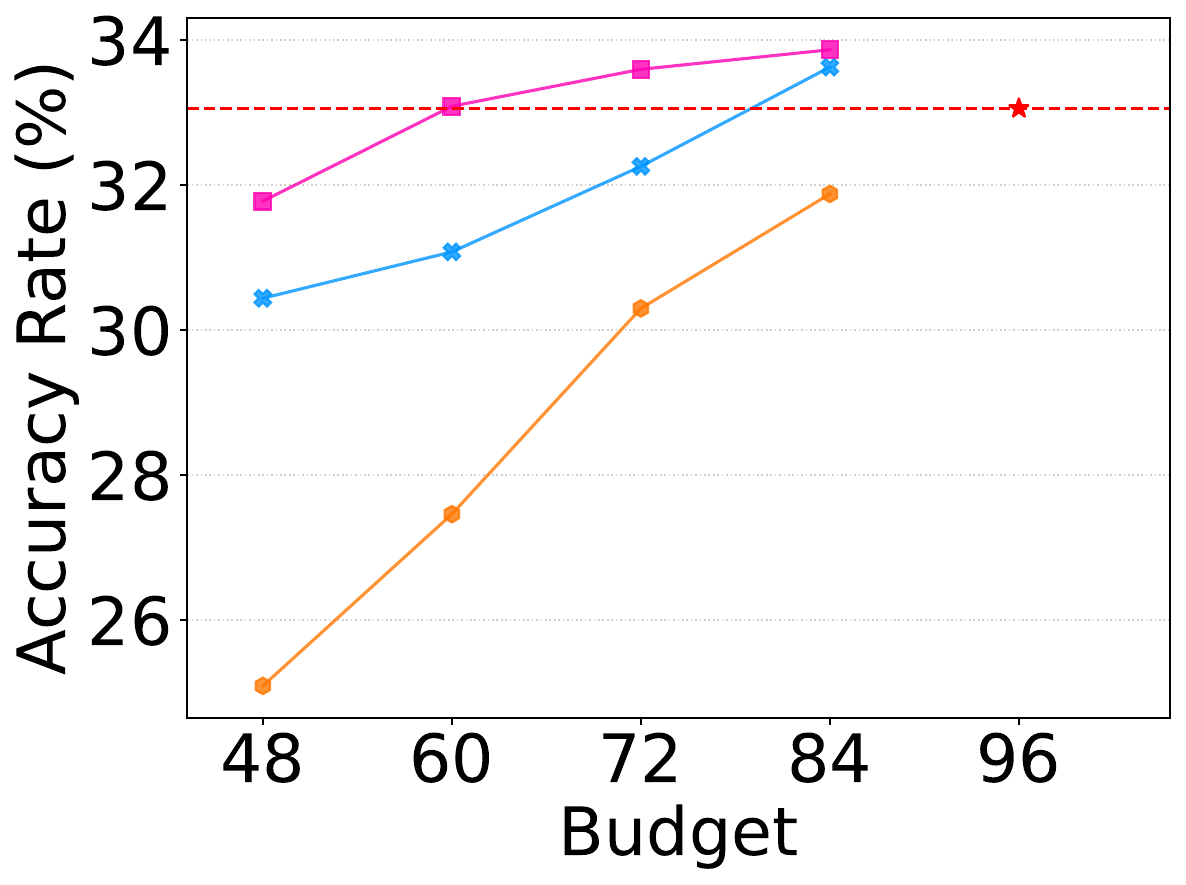}
  \vspace{-2mm}
  \small {\hspace{2em}(c)}
\end{minipage}

\vspace{2mm}

% ---------- (a) ----------
\begin{minipage}{0.32\linewidth}
  \centering
  \includegraphics[width=\linewidth]{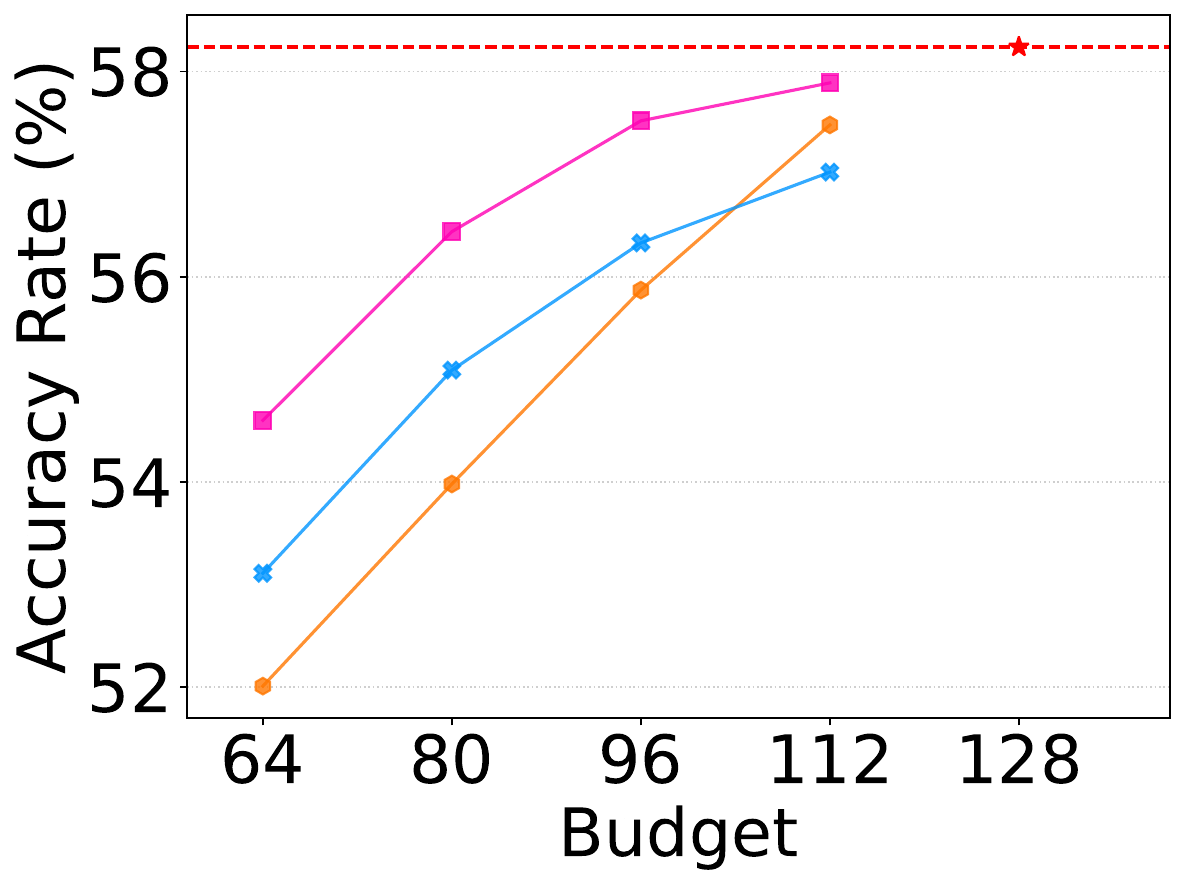}
  \vspace{-2mm}
  \small {\hspace{2em}(d)}
\end{minipage}\hfill
% ---------- (b) ----------
\begin{minipage}{0.32\linewidth}
  \centering
  \includegraphics[width=\linewidth]{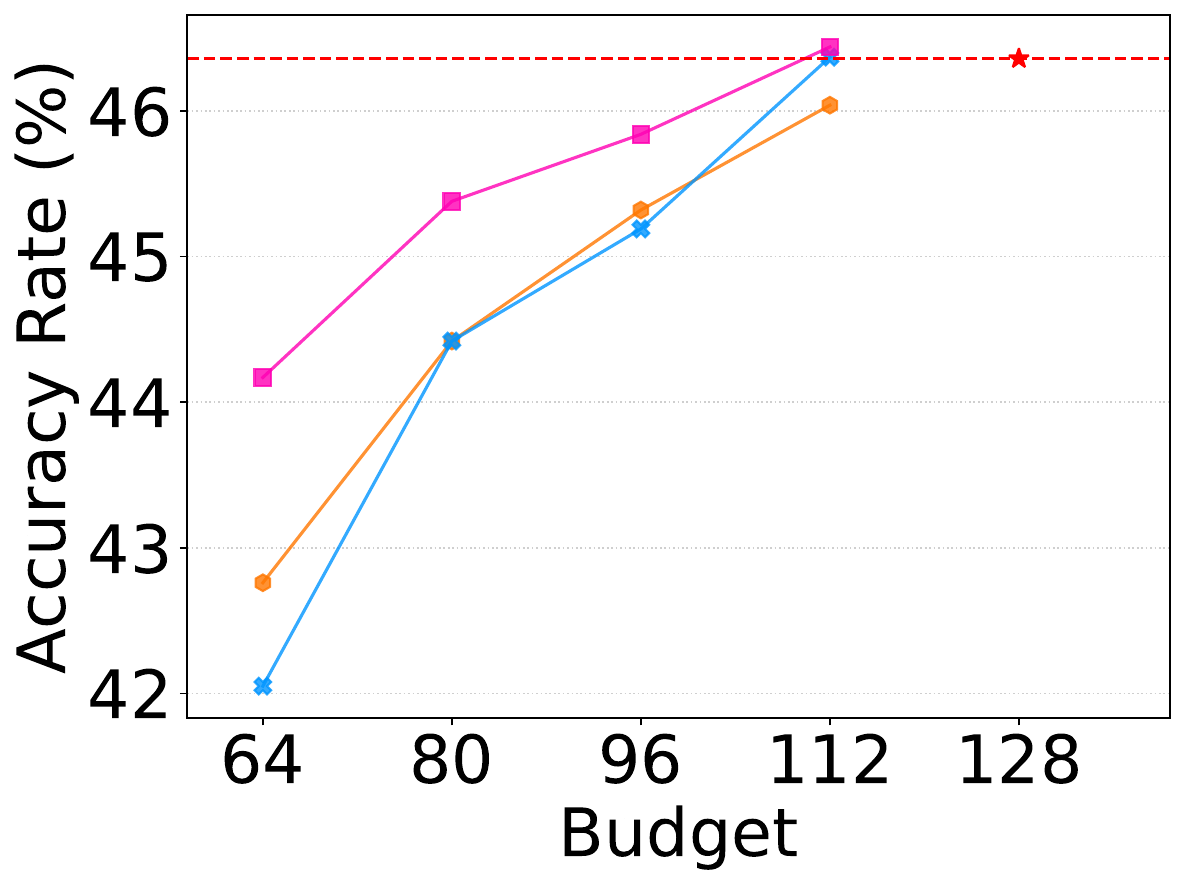}
  \vspace{-2mm}
  \small {\hspace{2em}(e)}
\end{minipage}\hfill
% ---------- (c) ----------
\begin{minipage}{0.32\linewidth}
  \centering
  \includegraphics[width=\linewidth]{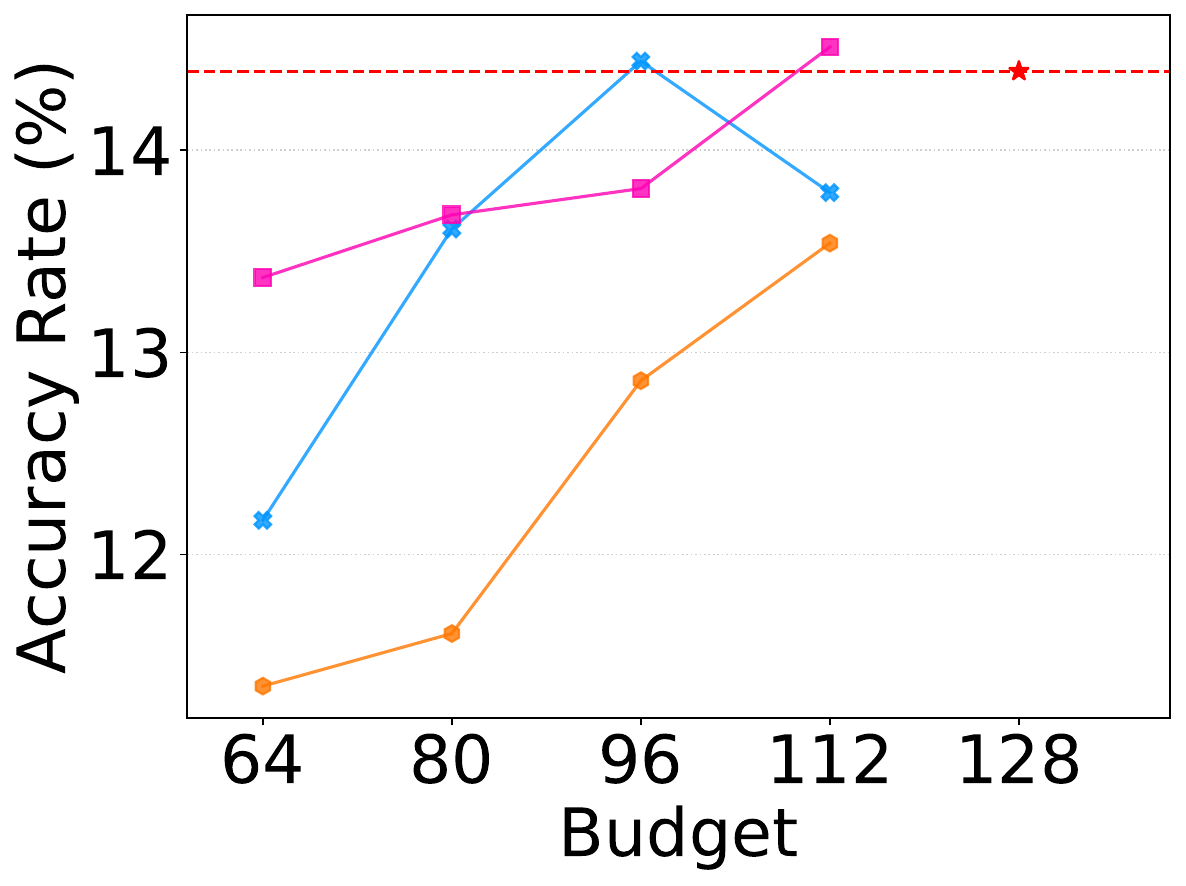}
  \vspace{-2mm}
  \small {\hspace{2em}(f)}
\end{minipage}

\caption{Ablation results of Alloc-T on Qwen1.5-MoE-A2.7B and OLMoE-1B-7B-0924 under varying global activation budgets.
(a–c) report results on Qwen1.5-MoE-A2.7B, while (d–f) correspond to OLMoE-1B-7B-0924.
For each model, (a,d) show NLU tasks, (b,e) Reasoning tasks, and (c,f) Math tasks.}
\label{fig:Qwen and OLMoE Alloc-T}
\end{figure*}

\begin{figure*}[t]
\centering
% ---------- (a) ----------
\hspace*{-0.8cm}  % 左移第一个子图，负值越大左移越多
\begin{minipage}{0.43\linewidth}
  \centering
  \includegraphics[width=\linewidth]{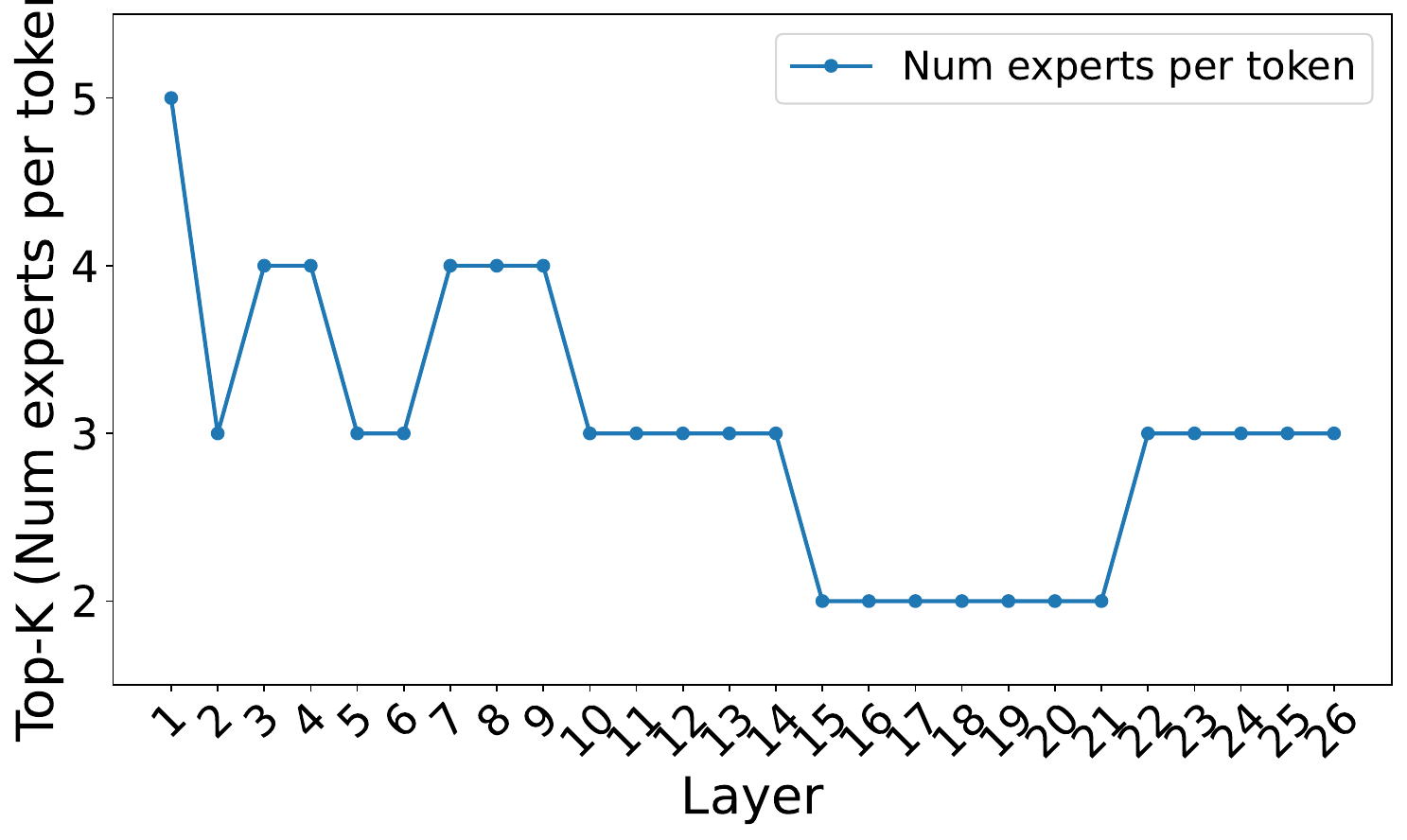}
  \vspace{-2mm}
  \small {\hspace{2em}(a)}
\end{minipage}
% ---------- (b) ----------
\hspace*{0.5cm}
\begin{minipage}{0.4\linewidth}
  \centering
  \includegraphics[width=\linewidth]{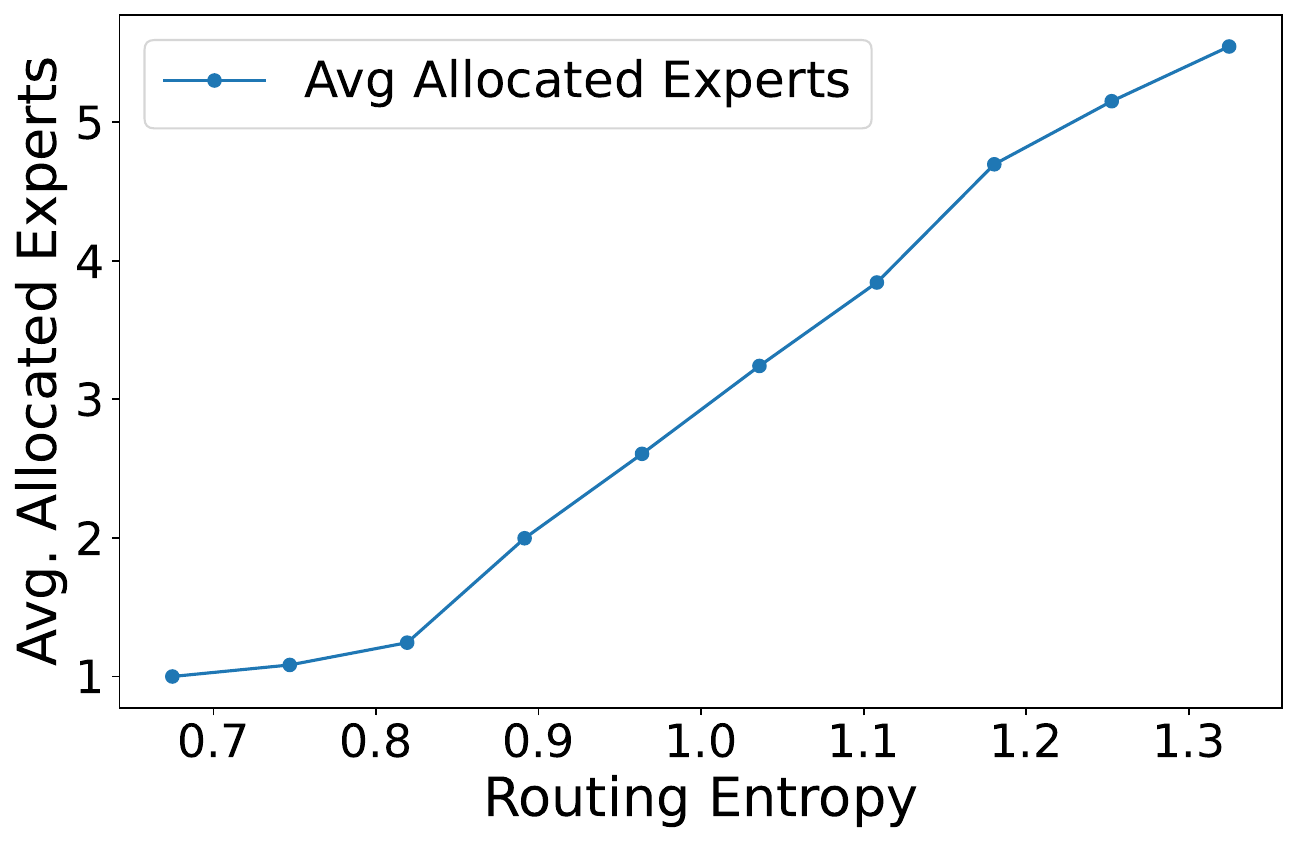}
  \vspace{-2mm}
  \small {\hspace{2em}(b)}
\end{minipage}
\caption{(a) Layer-wise allocation of Alloc-MoE on DeepSeek-V2-Lite with Budget = 78, demonstrating a clearly non-uniform distribution across layers. Earlier layers retain larger expert budgets, while deeper layers operate with fewer activated experts. (b) Entropy results on the 10-th MoE layer of DeepSeek-V2-Lite. Low-entropy (high-confidence) tokens retain fewer experts, whereas high-entropy (ambiguous) tokens retain more. Similar patterns are observed across layers.}
\label{fig:allocation_across_layers_and_tokens}
\end{figure*}

\end{document}